\documentclass[letterpaper, 10 pt, conference, twoside]{IEEEtran}  

\IEEEoverridecommandlockouts 

\usepackage[english]{babel}
\usepackage[nolist]{acronym}
\usepackage{hyperref}
\usepackage{multirow}
\usepackage[utf8]{inputenc}
\usepackage{soul}
\usepackage[most]{tcolorbox}

\makeatletter
\let\MYcaption\@makecaption
\makeatother

\usepackage[font=scriptsize]{subcaption}

\makeatletter
\let\@makecaption\MYcaption
\makeatother

\usepackage{array}
\newcommand{\PreserveBackslash}[1]{\let\temp=\\#1\let\\=\temp}
\newcolumntype{C}[1]{>{\PreserveBackslash\centering}p{#1}}
\newcolumntype{R}[1]{>{\PreserveBackslash\raggedleft}p{#1}}
\newcolumntype{L}[1]{>{\PreserveBackslash\raggedright}p{#1}}

\usepackage{cite} 

\usepackage{pifont}

\usepackage{dblfloatfix}
\usepackage{placeins}
\usepackage[export]{adjustbox}

\usepackage{booktabs}
\usepackage{enumitem}
\newlist{questions}{enumerate}{2}
\setlist[questions,1]{label=\textbf{RQ\arabic*.}, ref=RQ\arabic*}
\setlist[questions,2]{label=(\alph*), ref=\thequestionsi(\alph*)}

\usepackage{makecell}

\usepackage{siunitx}
\usepackage{graphics} 
\usepackage{graphicx} 

\usepackage{tikz}
\pgfdeclarelayer{foreground}
\pgfsetlayers{background, main, foreground}
\usetikzlibrary{quotes, angles, backgrounds, arrows, automata, shapes, positioning, calc, through, spy, arrows.meta, 3d, perspective}

\tikzset{
    imglabel/.style={
      rectangle,
      inner sep=2pt,
      text=black,
      minimum height=1em,
      text centered,
      fill=white,
      fill opacity=1.0,
      text opacity=1,
      anchor=south west,
    },
  }

\tikzset{
	state/.style={
		rectangle,
		draw=black, very thick,
		minimum height=1.0em,
		text centered,
	},
}

\usepackage{algorithm, algorithmic} 

\usepackage{amsmath} 
\usepackage{amssymb}  
\usepackage{bm}

\usepackage{array} 
\usepackage{wrapfig} 

\newcommand{\mJ}{\mathbf{J}} 
\newcommand{\vp}{\mathbf{p}} 
\newcommand{\vv}{\mathbf{v}} 
\newcommand{\vomega}{\boldsymbol{\omega}} 
\newcommand{\vfa}{\mathbf{f}_a} 
\newcommand{\vtaua}{\boldsymbol{\tau}_a} 

\newcommand{\tflag}[1]{}
\newcommand{\tfflag}[1]{}

\usepackage{colortbl} 

\newcommand{\cmark}{{\noindent\color{green}{\ding{51}}}}%
\newcommand{\xmark}{{\noindent\color{red}{\ding{55}}}}%
\newcommand{\amark}{{\noindent\color{orange}{\ding{81}}}}%

\newcounter{dynamics}

\usepackage{scalerel}
\usetikzlibrary{svg.path}
\definecolor{orcidlogocol}{HTML}{A6CE39}
\tikzset{
	orcidlogo/.pic={
		\fill[orcidlogocol] 
		svg{M256,128c0,70.7-57.3,128-128,128C57.3,256,0,198.7,0,128C0,57.3,57.3,0,128,0C198.7,0,256,57.3,256,128z};
		\fill[white] svg{M86.3,186.2H70.9V79.1h15.4v48.4V186.2z}
		svg{M108.9,79.1h41.6c39.6,0,57,28.3,57,53.6c0,27.5-21.5,53.6-56.8,53.6h-41.8V79.1z 
		M124.3,172.4h24.5c34.9,0,42.9-26.5,42.9-39.7c0-21.5-13.7-39.7-43.7-39.7h-23.7V172.4z}
		svg{M88.7,56.8c0,5.5-4.5,10.1-10.1,10.1c-5.6,0-10.1-4.6-10.1-10.1c0-5.6,4.5-10.1,10.1-10.1C84.2,46.7,88.7,51.3,88.7,56.8z};
	}
}
\newcommand\orcidicon[1]{\href{https://orcid.org/#1}{\mbox{\scalerel*{
				\begin{tikzpicture}[yscale=-1,transform shape]
				\pic{orcidlogo};
				\end{tikzpicture}
			}{|}}}}

\title{\LARGE \bf
    Survey of Simulators for Aerial Robots \\\large{An Overview and In-Depth Systematic Comparisons}\vspace*{-2mm}
} 

\author{Cora A. Dimmig$^{1,2}$,  Giuseppe Silano$^{3}$, Kimberly McGuire$^{4}$, Chiara Gabellieri$^{5}$, Wolfgang Hönig$^{6}$, \\ Joseph Moore$^{1,2}$, and Marin Kobilarov$^{1}$
\vspace*{-3mm}
    \thanks{$^{1}$Department of Mechanical Engineering, Johns Hopkins University, Baltimore, Maryland, USA (emails: {\tt\small \{cdimmig, marin\}@jhu.edu}).}
    \thanks{$^{2}$Johns Hopkins University Applied Physics Laboratory, Laurel, Maryland, USA (email: {\tt\small joseph.moore@jhuapl.edu}).}
    \thanks{$^{3}$Department of Cybernetics, Czech Technical University in Prague, Prague, Czech Republic (email:  {\tt\small giuseppe.silano@fel.cvut.cz}).}
    \thanks{$^{4}$Bitcraze A.B., Malmo, Sweden (email: {\tt\small kimberly@bitcraze.io}).}
    \thanks{$^{5}$Robotics and Mechatronics (RaM) Group, University of Twente, Enschede, The Netherlands (email: {\tt\small c.gabellieri@utwente.nl}).}
    \thanks{$^{6}$Intelligent Multi-Robot Coordination Lab, Technische Universität (TU) Berlin, Germany (email: {\tt\small hoenig@tu-berlin.de}).}
    \thanks{This work was partially funded by the National Science Foundation grant no. 1925189, the EU's MSCA FLYFLIC grant no. 101059875, the EU's H2020 AERIAL-CORE grant no. 871479, and the Deutsche Forschungsgemeinschaft (DFG, German Research Foundation) - 448549715.}
    
}%

\begin{document}

\maketitle
\thispagestyle{empty}
\pagestyle{empty}


\begin{acronym}
    \acro{AR}[AR]{Aerial Vehicle}
    \acro{API}[API]{Application Program Interface}
    \acro{CoM}[CoM]{Center of Mass}
    \acro{DOPE}[DOPE]{Deep Object Pose Estimation}
    \acro{DoF}[DoF]{Degree-of-Freedom}
    \acro{HITL}{Hardware-In-The-Loop}
    \acro{HSI}{Human-Swarm Interaction}
    \acro{ICRA}{International Conference on Robotics and Automation}
    \acro{ML}[ML]{Machine Learning}
    \acro{RL}[RL]{Reinforcement Learning}
    \acro{ROS}[ROS]{Robot Operating System}
    \acro{SITL}{Software-In-The-Loop}
    \acro{UAV}[UAV]{Uncrewed Aerial Vehicle}
    \acro{wrt}[w.r.t.]{with respect to}
    \acro{VTOL}[VTOL]{Vertical Take-Off and Landing}
    \acro{RC}[RC]{Radio Controlled}
    \acro{ODE}[ODE]{Open Dynamics Engine}
    \acro{OS}[OS]{Operating System}
\end{acronym}


\begin{figure*}[t]
    \begin{center}
    \begin{adjustbox}{minipage=\linewidth,scale=0.95}
	\begin{subfigure}[t]{0.4\textwidth}
    	\centering
    	\includegraphics[height=0.17\textheight]{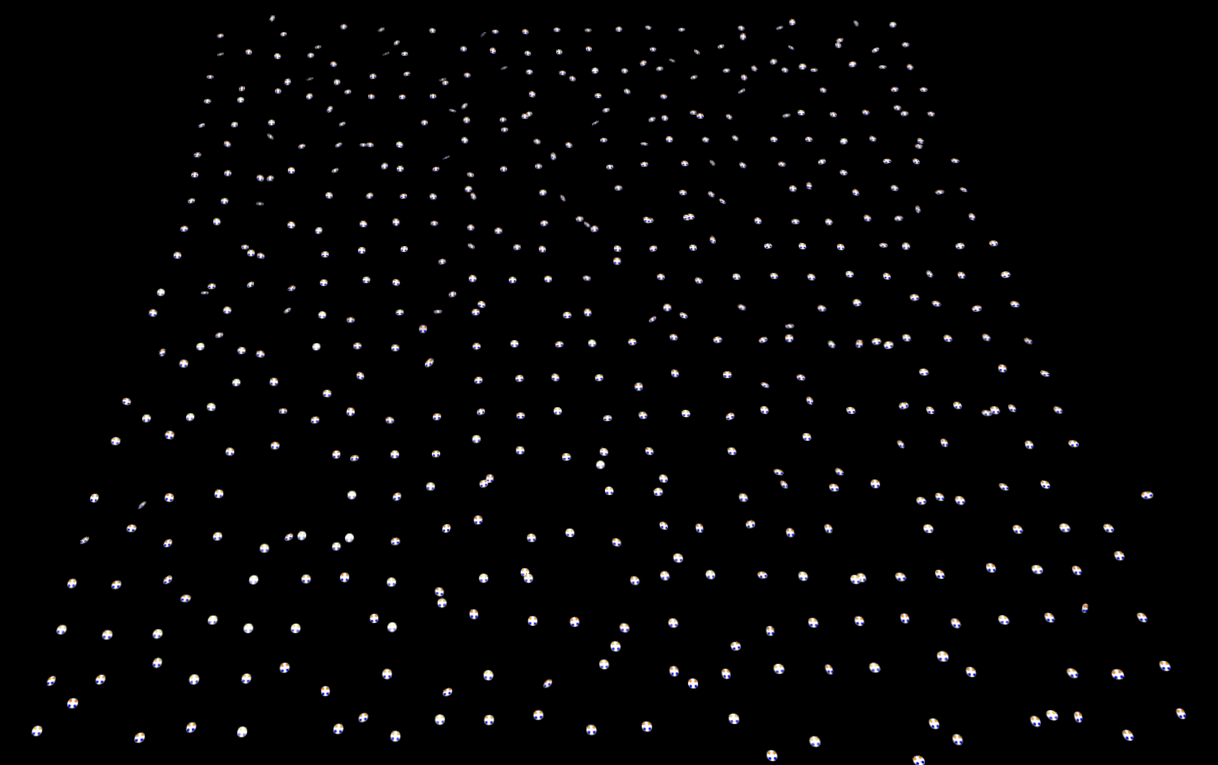}
    	\captionsetup{font=scriptsize}
            \vspace{-0.40em}
            \caption{Isaac Gym (Aerial Gym)}
    	\label{fig:isaac_gym}
	\end{subfigure}
	\begin{subfigure}[t]{0.33\textwidth}
    	\centering
    	\includegraphics[height=0.17\textheight]{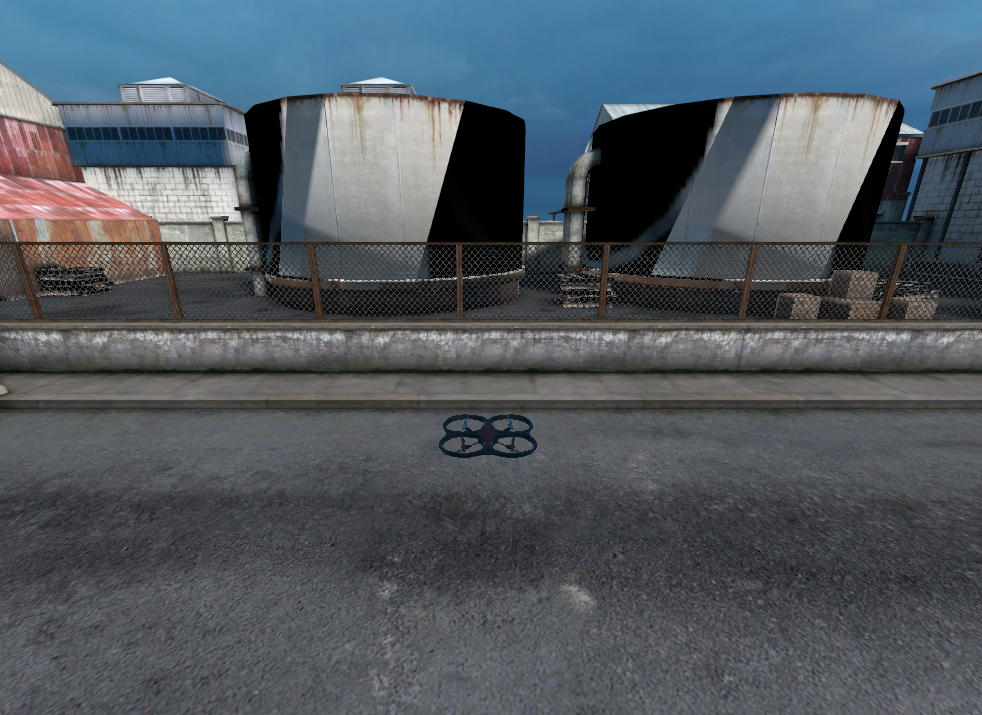}
    	\captionsetup{font=scriptsize}
            \vspace{-0.40em}
            \caption{Flightmare}
    	\label{fig:flightmare}
	\end{subfigure}
	\begin{subfigure}[t]{0.23\textwidth}
    	\centering
    	\includegraphics[height=0.17\textheight]{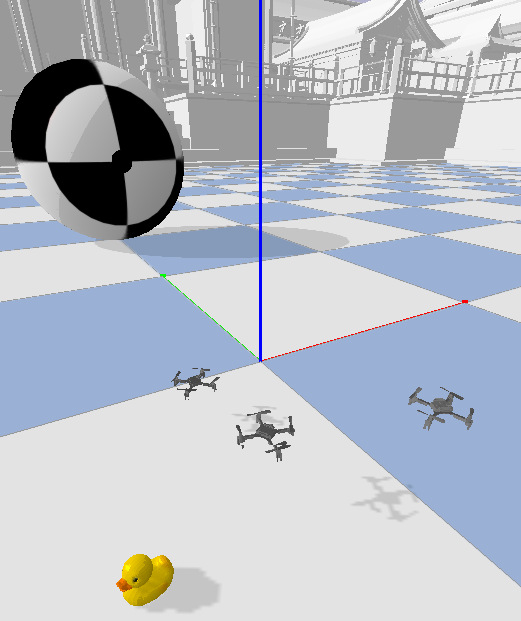}
    	\captionsetup{font=scriptsize}
            \vspace{-0.40em}
            \caption{gym-pybullet-drones}
    	\label{fig:gym-pybullet-drones}
	\end{subfigure}

	\begin{subfigure}[t]{0.25\textwidth}
		\centering
    	\includegraphics[height=0.17\textheight]{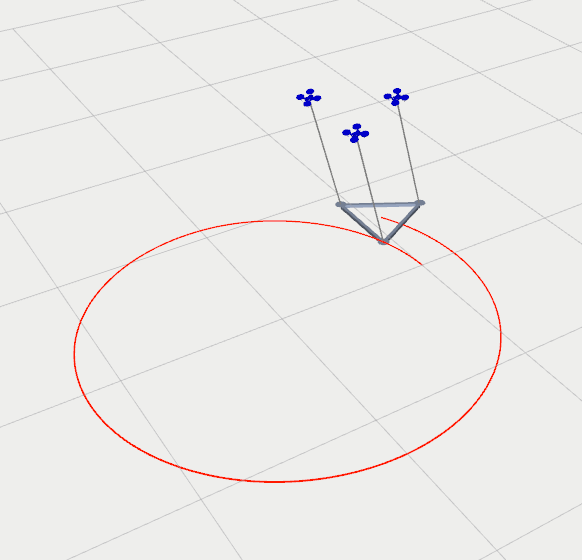}
    	\captionsetup{font=scriptsize}
            \vspace{-0.40em}
            \caption{RotorTM}
    	\label{fig:RotorTM}
	\end{subfigure}
	\begin{subfigure}[t]{0.31\textwidth}
		  \centering
		  \includegraphics[height=0.17\textheight]{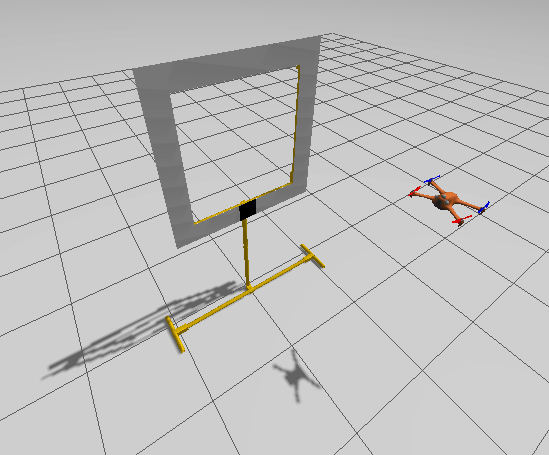}
		  \captionsetup{font=scriptsize}
            \vspace{-0.40em}
            \caption{Aerostack2}
		\label{fig:aerostack2}
	\end{subfigure}
	\begin{subfigure}[t]{0.4\textwidth}
		  \centering
		  \includegraphics[height=0.17\textheight]{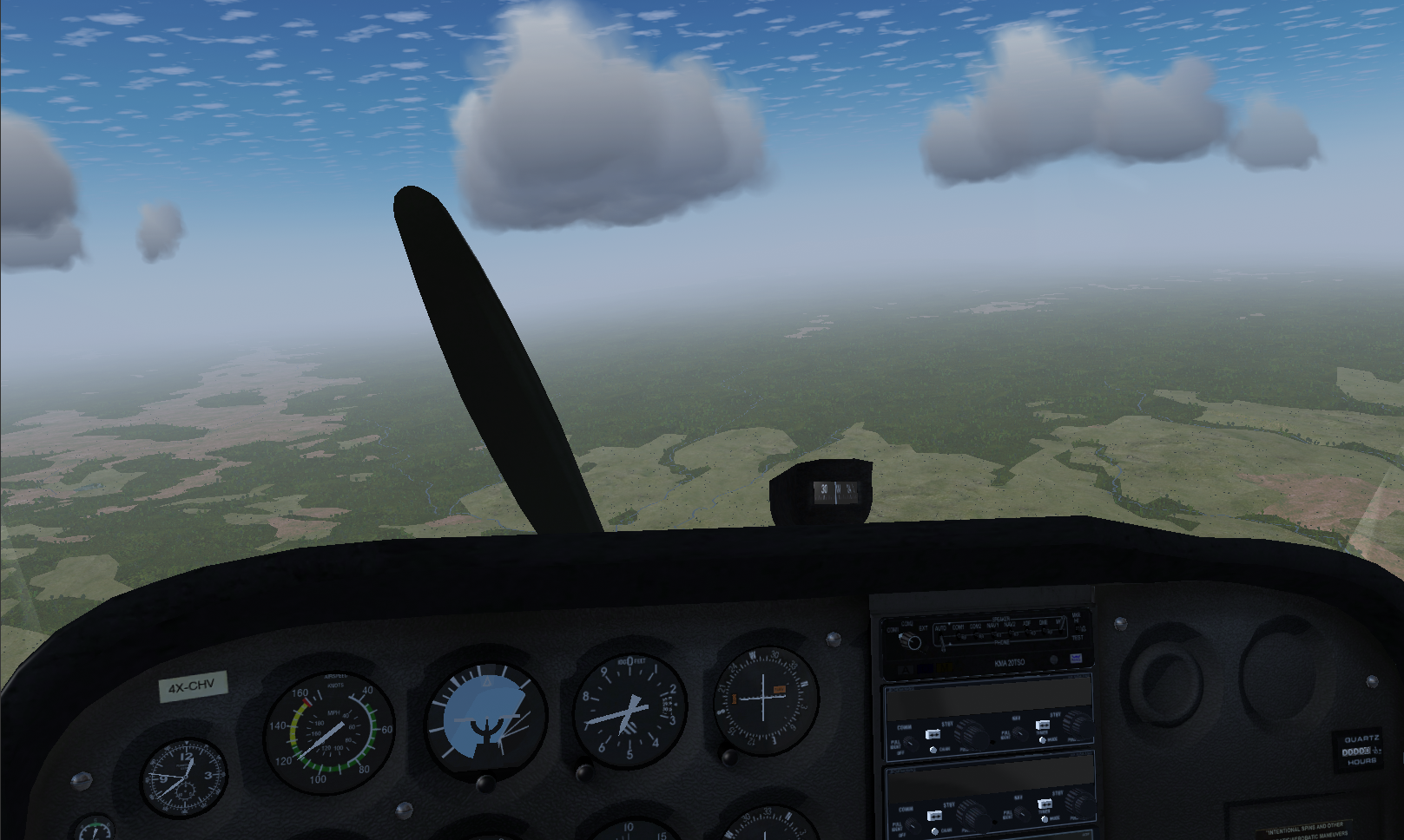}
		  \captionsetup{font=scriptsize}
            \vspace{-0.40em}
            \caption{FlightGear}
		  \label{fig:flightgear}
	\end{subfigure}

    \end{adjustbox}
    \vspace*{-0.5em}
    \caption{Examples of simulation environments: (a) 512 quadrotors in Aerial Gym, (b) quadrotor in Flightmare, (c) three quadrotors in gym-pybullet-drones, (d) three quadrotors with a triangular payload in RotorTM, (e) quadrotor with a racing gate in Aerostack2, and (f) Cessna 172P cockpit view in FlightGear.}
    \label{fig:simulators}
    \end{center}
    \vspace*{-2.0em}
\end{figure*}


\begin{abstract}

\ac{UAV} research faces challenges with safety, scalability, costs, and ecological impact when conducting hardware testing. 
High-fidelity simulators offer a vital solution by replicating real-world conditions to enable the development and evaluation of novel perception and control algorithms. 
However, the large number of available simulators poses a significant challenge for researchers to determine which simulator best suits their specific use-case, based on each simulator's limitations and customization readiness. 
In this paper we present an overview of 44 \ac{UAV} simulators, including in-depth, systematic comparisons for 14 of the simulators. Additionally, we present a set of decision factors for selection of simulators, aiming to enhance the efficiency and safety of research endeavors.

\end{abstract}


\section{Introduction}
\label{sec:introduction}

\acfp{UAV} are being widely adopted for a variety of use cases and industries, such as agriculture, inspection, mapping, and search and rescue~\cite{Leutenegger2016}. In particular, aerial manipulation and human-robot interaction applications have been on the rise, including tasks such as parcel delivery, 
sample collection, and collaborative robot operations~\cite{OlleroFranchi2022Past, KollingHMS2016}. 

Testing experimental algorithms directly on hardware can pose significant risks, as unexpected behaviors may emerge. Moreover, crashes can incur substantial costs, disrupt development schedules, and contribute to environmental harm due to the frequent replacement of damaged vehicle components. Additionally, in the context of the increasing adoption of machine learning-based techniques, collecting data from hardware can prove highly inefficient and often impracticable. 
Hence, 
a dependable, fast, precise, and realistic \ac{UAV} simulator is essential to facilitate rapid advancements in this field. 
Due to the rise of high-fidelity simulators, results from simulation can often be efficiently transferred to hardware, however challenges may arise in domains with unmodeled effects (e.g. agile flight, close-proximity flight, \acp{UAV} with manipulators). 

In this work, we analyze some of the prominent~\ac{UAV} simulators and discuss key selection criteria and decision factors to consider when choosing a simulator. 
To illustrate the breadth of existing simulators, we include simulators that are widely used (as high adoption rates often result in more testing and community input) and simulators specialized for popular research areas.
Figure~\ref{fig:simulators} demonstrates the large spectrum of simulators and their use cases.
This research builds upon discussions and contributions made during the workshop titled ``\textit{The Role of Robotics Simulators for Unmanned Aerial Vehicles}'' at the 2023~\ac{ICRA} in London, UK \cite{WorkshopLink}. Specifically, our work addresses the following question:
\vspace*{-0.5em}
\begin{table}[ht]
    \centering
    \begin{tabular}{r p{6.58cm}}
         \toprule
         \normalsize{\textbf{Question:}} & \normalsize{What criteria and considerations should guide the selection and customization of a simulator to optimize its suitability for a specific application, while also understanding its limitations?}  \\
         \bottomrule
    \end{tabular}
\end{table}
\vspace*{-0.65em}

There are a few existing survey papers focusing on simulators and their role in robotics~\cite{Liu2021ARCRAS}.
For example, a recent survey~\cite{CollinsHoward2021Review} analyzes a wide range of application areas, including aerial vehicles, 
and compares features across diverse domains. 
Regrettably, not every simulator readily supports \acp{UAV}. 
Dynamics considerations for manipulators or ground vehicles can substantially differ from those of aerial vehicles, especially in research that aims to account for aerodynamic effects. In \cite{SaundersLi2022Autonomous}, the authors examine various considerations for aerial delivery vehicles, including simulator selection. 
In \cite{MairajJavaid2019Application}, the authors analyze simulators specific to aerial vehicles, including some less commonly used simulators, and discuss criteria for simulator selection.

We believe that the ensuing discussion and difficulty in selecting a simulator requires a more focused survey paper covering an expanded list of \ac{UAV} simulators. We provide readers with valuable insights based on our experiences in international robotics competitions, innovative research projects, and real-world applications, contemplate the future of simulation tools, and provide consolidated information for readers to explore effective solutions for their intended applications.



\begin{tcolorbox}[breakable]
\section*{UAV Dynamics Background}
\refstepcounter{dynamics}
\label{sec:background}
\small

This section delves into fundamental concepts crucial for a comprehensive understanding of~\ac{UAV} dynamics. We focus on \acp{UAV} without morphing capabilities (i.e. the ability for a vehicle to change its shape).
We express variables and parameters in Table~\ref{tab:uav_parameters} and vehicle schematics in Figure~\ref{fig:schematicRepresentationMultirotor}.



\subsection{Multirotors}
\label{sec:multirotors}

\textbf{Basic}: The multirotor's dynamics are modeled as a 6-\ac{DoF} floating rigid body using Newton-Euler formalism with squared motor speed inputs~\cite{Leutenegger2016}. 
The motors exert forces and torques on the~\ac{CoM}.

\quad\textit{State:} $\mathbf{x}=(\vp, \mathbf{q}, \vv, \vomega)^\top$,
\vspace*{0.6mm}

\quad\textit{Control Inputs:} $\mathbf{u}_{\Omega}=(\Omega_1,\hdots,\Omega_n)^\top$,
\vspace*{0.6mm}

\quad\textit{Forces:} $\mathbf{f} = \sum_i^n c_{f_i} \Omega_i \mathbf{z}_{\Omega_i} = \mathbf{F} \mathbf{u}_{\Omega}$,
\vspace*{0.6mm}

\quad\textit{Torques:} $\bm{\tau} = \sum_i^n ( c_{f_i} \mathbf{p}_{\Omega_i} \times \mathbf{z}_{\Omega_i} + c_{\tau_i} \mathbf{z}_{\Omega_i} ) {\Omega}_i = \mathbf{M} \mathbf{u}_{\Omega}$,
\vspace*{0.6mm}

\quad\textit{Dynamics:}
\vspace*{-1.8mm}
\begin{equation}
\label{eq:multirotorDynamics}
\resizebox{0.87\hsize}{!}{$
   \begin{array}{l}
       \Dot{\mathbf{p}} = \mathbf{v}, \quad\quad\quad\quad~~\,
       m\dot{\mathbf{v}} = m\mathbf{g} + \mathbf{R}(\mathbf{q}) \mathbf{F} \mathbf{u}_{\Omega} +\mathbf{f}_a,\\
       \Dot{\mathbf{q}} = \cfrac{1}{2} \, \mathbf{q} \circ \begin{bmatrix} 0 \\ \bm{\omega} \end{bmatrix}, ~~\;\;
       \mathbf{J}\Dot{\bm{\omega}} = -\bm{\omega} \times \mathbf{J}\bm{\omega}  + \mathbf{M} \mathbf{u}_{\Omega} + \bm{\tau}_a,
   \end{array}
   $}
\vspace*{-1mm}
\end{equation}
where $\circ$ and $\times$ represent quaternion and vector cross products, respectively.
The model can be extended for \acp{UAV} with tilting propellers by introducing a control variable $\mathbf{u}_w$ for real-time adjustment of $\mathbf{F}(\mathbf{u}_w)$ and $\mathbf{M}(\mathbf{u}_w)$ matrices \cite{OlleroFranchi2022Past}. 

\textbf{Drag}: At high speeds, multirotors experience aerodynamic drag forces 
and torques, 
often considered disturbances 
proportional to the velocity ($\vfa \propto \vv$ and $\vtaua \propto \bm{\omega}$)~\cite{faesslerDifferentialFlatnessQuadrotor2018}. 

\textbf{Wind}: Wind is typically modeled by a spatio-temporal-varying external force $\mathbf{f}_a(\mathbf{p}_w, t)$, where $\mathbf{p}_w$ and $t $ represent position and time, respectively~\cite{Leutenegger2016}.

\textbf{Interactions}: In close-proximity flight, multirotors experience aerodynamic interaction forces, often modeled as a learned function $\mathbf{f}_a$ based on relative neighbor poses~\cite{NeuralSwarm2}.



\subsection{Helicopters}
\label{sec:helicopters}

\textbf{Basic}: 
Helicopters 
have similar dynamics to multirotors,
mainly differing in force $\mathbf{F}$ and torque $\mathbf{M}$ allocation matrices, which are two-dimensional and potentially not fixed~\cite{Leutenegger2016}.

    \vspace*{-1.0em}
    \begin{center}
    \scalebox{0.47}{
    \definecolor{darkgreen}{rgb}{0.272, 0.50, 0.376}
    \definecolor{lightgreen}{rgb}{0.585, 0.82, 0.647}
    \begin{tikzpicture}[scale=2.0, line cap=round, line join=round, >=Triangle]
        \draw[color=black!60, rotate around={-110:(0,0)}, fill=darkgreen!80, line width=1pt] (0,0) ellipse (0.95cm and 1.50cm);
        \draw[color=black!60, rotate around={-20:(0,0)}, fill=lightgreen!80, dashed, line width=1.5pt] (0,0) ellipse (1.5cm and 0.75cm);

        \fill (0,0) -- ++(0.2em,0) arc [start angle=0, end angle=90, radius=0.2em] -- ++(0,-0.4em) arc [start angle=270, end angle=180, radius=0.2em];
       \draw (0,0) [radius=0.2em] circle;

        \draw (-0.05,0) node[left]{$O_B$}; 
        \draw [->] (0,0) -- ({0.5*cos(65)},{0.5*sin(65)}) node[left]{$\mathbf{z}_B$}; 
        \draw [->] (0,0) -- ({0.5*cos(20)},{0.5*sin(20)}) node[above]{$\mathbf{y}_B$}; 
        \draw [->] (0,0) -- ({0.5*cos(-25)},{0.5*sin(-25)}) node[below]{$\mathbf{x}_B$}; 

        \draw (-1.5,-1.5) node[left]{$O_W$}; 
        \draw [->] (-1.5,-1.5) -- (-1.5,-1.0) node[left]{$\mathbf{z}_W$}; 
        \draw [->] (-1.5,-1.5) -- (-1.15,-1.25) node[right]{$\mathbf{y}_W$}; 
        \draw [->] (-1.5,-1.5) -- (-1.0,-1.5) node[below]{$\mathbf{x}_W$}; 

        \draw[->, red] (-1.5,-1.5) -- node[above]{$\mathbf{p}$} (0,0);
        \draw[->, red] (-1.0,-1.75) to [out=-20, in=-70] node[left]{$\mathbf{R}$} (0.15, -0.15);

        \tikzset
        {%
          pics/cylinder/.style n args={3}{
            code={%
              \draw[pic actions] (135:#1) arc (135:315:#1) --++ (0,0,#2) arc (315:135:#1) -- cycle;
              \draw[pic actions] (0,0,#2) circle (#1);
              \foreach\z in {0,1}
              {
                \begin{scope}[canvas is xy plane at z=\z*\h]
                  \coordinate (-cen\z) at       (0,0);
                  \coordinate (-ESE\z) at    (-#3:#1);
                  \coordinate (-ENE\z) at     (#3:#1);
                  \coordinate (-NNE\z) at  (90-#3:#1);
                  \coordinate (-NNW\z) at  (90+#3:#1);
                  \coordinate (-WNW\z) at (180-#3:#1);
                  \coordinate (-WSW\z) at (180+#3:#1);
                  \coordinate (-SSW\z) at (270-#3:#1);
                  \coordinate (-SSE\z) at (270+#3:#1);
                \end{scope}
              }
            }},
        }
        \def\l{2}    
        \def\R{1}    
        \def\r{0.15} 
        \def\h{0.3}  
        \pic[fill=gray!30, rotate=-15, shift={(0,0,-\h)}] at (1.25,0.10) {cylinder={\r}{2*\h}{45}};
        \draw[->] (1.25,0.1225) -- (1.55,0.30) node[above]{$\mathbf{z}_P$};
        \begin{scope}[shift={(1.405,0.25)},rotate around z=120,canvas is xy plane at z=\h]
          \draw[fill=gray!30] (0.0,0) sin  (0.25,0.1) cos  (0.65,0) sin  (0.25,-0.1) cos (0,0)sin (-0.25,0.1) cos (-0.65,0) sin (-0.25,-0.1) cos (0,0);
          \fill (0,0) circle (0.05);
          \draw[latex-] (0.115,-0.175) arc (020:160:0.1) node [below] {$\tau$};
        \end{scope}
        \pic[fill=gray!30, rotate=175, shift={(0,0,-\h)}] at (-1.25,-0.10) {cylinder={\r}{2*\h}{45}};
        \draw[->] (-1.275,-0.11) -- (-1.545,-0.345) node[above]{$\mathbf{z}_P$};
        \begin{scope}[shift={(-1.18,-0.025)},rotate around z=130,canvas is xy plane at z=\h]
          \draw[fill=gray!30] (0.0,0) sin  (0.25,0.1) cos  (0.65,0) sin  (0.25,-0.1) cos (0,0)sin (-0.25,0.1) cos (-0.65,0) sin (-0.25,-0.1) cos (0,0);
          \draw[-latex] (0.1,0.075) arc (020:160:0.1) node [below] {$\tau$};
          \fill (0,0) circle (0.05);
        \end{scope}
        \pic[fill=gray!30, rotate=70, shift={(0,0,-\h)}] at (-0.835,1.035) {cylinder={\r}{2*\h}{45}};
        \draw[->] (-0.845,1.045) -- (-0.965,1.335) node[right]{$\mathbf{z}_P$};
        \begin{scope}[shift={(-0.75,1.20)},rotate around z=20,canvas is xy plane at z=\h]
          \draw[fill=gray!30] (0.0,0) sin  (0.25,0.1) cos  (0.65,0) sin  (0.25,-0.1) cos (0,0)sin (-0.25,0.1) cos (-0.65,0) sin (-0.25,-0.1) cos (0,0);
          \draw[latex-] (0.075,0.075) arc (00:140:0.1) node [left] {$\tau$};
          \fill (0,0) circle (0.05);
        \end{scope}
        \pic[fill=gray!30, rotate=240, shift={(0,0,-\h)}] at (0.85,-0.80) {cylinder={\r}{2*\h}{45}};
        \draw[->] (0.865,-0.85) -- (0.95,-1.10) node[right]{$\mathbf{z}_P$};
        \begin{scope}[shift={(0.985,-0.74)},rotate around z=15,canvas is xy plane at z=\h]
          \draw[fill=gray!30] (0.0,0) sin  (0.25,0.1) cos  (0.65,0) sin  (0.25,-0.1) cos (0,0)sin (-0.25,0.1) cos (-0.65,0) sin (-0.25,-0.1) cos (0,0);
          \draw[-latex] (0.1,-0.2) arc (00:140:0.1) node [below left] {$\tau$};
          \fill (0,0) circle (0.05);
        \end{scope}
    \end{tikzpicture}
    }
    \scalebox{0.47}{
        \centering
        \begin{tikzpicture}[scale=2.0, line cap=round, line join=round, >=Triangle]
            \node at (0,1.0) [text centered]{\includegraphics[scale=0.5]{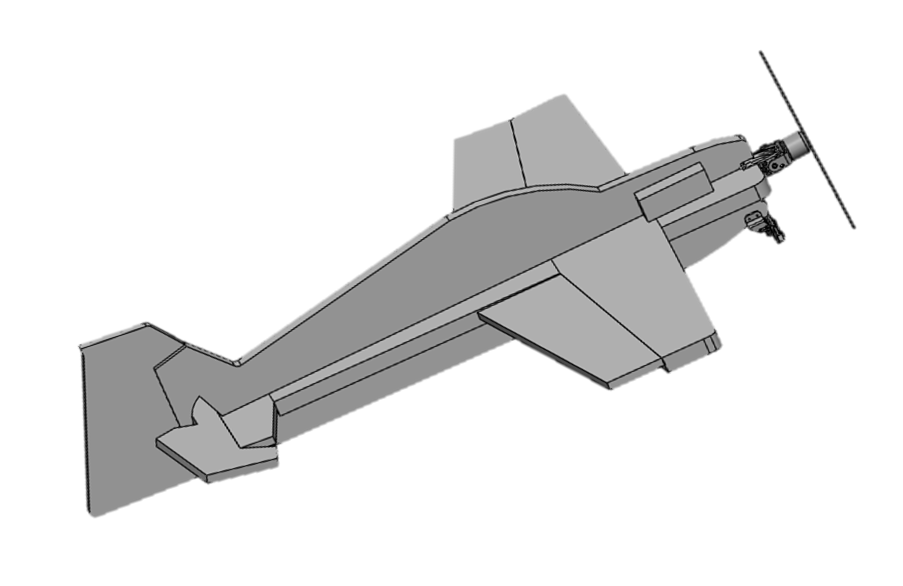}};

            \draw (-1.5,-1.0) node[left]{$O_W$}; 
            \draw [->] (-1.5,-1.0) -- (-1.5,-0.5) node[left]{$\mathbf{z}_W$}; 
            \draw [->] (-1.5,-1.0) -- (-1.15,-0.75) node[right]{$\mathbf{y}_W$}; 
            \draw [->] (-1.5,-1.0) -- (-1.0,-1.0) node[below]{$\mathbf{x}_W$}; 

            \draw (0.55,1.2) node[left]{$O_B$}; 
            \draw [line width=0.25mm] [->] (0.55,1.2) -- (0.55,1.5) node[above right]{$\mathbf{z}_B$}; 
            \draw [line width=0.25mm] [->] (0.55,1.2) -- (0.44,1.44) node[below left]{$\mathbf{y}_B$}; 
            \draw [line width=0.25mm] [->] (0.55,1.2) -- (0.8,1.32) node[below]{$\mathbf{x}_B$}; 

            \fill (0.55,1.2) -- ++(0.2em,0) arc [start angle=0, end angle=90, radius=0.2em] -- ++(0,-0.4em) arc [start angle=270, end angle=180, radius=0.2em];
            \draw (0.55,1.2) [radius=0.2em] circle;

            \draw[->, red, line width=0.5pt] (-1.5,-1.0) -- node[above]{$\mathbf{p}$} (0.55,1.2);
            \draw[->, red] (-1.0,-1.25) to [out=-20, in=-70] node[left]{$\mathbf{R}$} (0.55,1.1);

            \draw [dashed, ->] (-1.35,0.5) node[above]{$\mathbf{f}_{s_r}$}  -- (-1.35,0.25);
            \draw [dashed, <-] (-1.04,0.5) node[above]{$\mathbf{f}_{s_e}$} -- (-1.02,0.3);
            \draw [->] (-0.2,0.7) -- node[below left]{$\mathbf{g}$}(-0.2,0.45);
            \draw [dashed, <-] (0.12,1.67) node[above]{$\mathbf{f}_{s_{a_2}}$} -- (0.15,1.42);
            \draw [dashed, <-] (0.37,1.75) node[above]{$\mathbf{f}_{s_{w_2}}$} -- (0.45,1.50);
            \draw [dashed, <-] (0.47,1.05) -- (0.52,0.85);
            \node at (0.52,0.83) [text centered]{$\mathbf{f}_{s_{a_1}}$};
            \draw [dotted, line width=0.5pt] (0.52,0.85) -- (0.55,1.2) node[below right]{$\mathbf{l}_{s_{a_1}}$};
            \draw [dashed, <-] (0.84,1.15) -- (0.9,0.95) node[right]{$\mathbf{f}_{s_{w_1}}$};
            \draw [dashed, ->] (1.5,1.65) -- (1.75,1.775) node[below]{$\mathbf{f}_t$};
        \end{tikzpicture}
    }
    \end{center}
    \vspace*{-2.0em}
    \captionof{figure}{Schematic representations of a multirotor and fixed-wing with global frames $O_W$ and body frames $O_B$.}
    \label{fig:schematicRepresentationMultirotor}
    \vspace*{0.0em}
{
    \begin{center}
\captionof{table}{\ac{UAV} Dynamics Variables}
    \vspace*{-4.5mm}
    \label{tab:uav_parameters}
    \footnotesize
    \renewcommand{\arraystretch}{1.3}
    \renewcommand{\tabcolsep}{1.4mm}
    \begin{tabular}{c | c | L{6cm} }
        \hline
        \textbf{Var.} & \textbf{Set} & \textbf{Description} \\\hline\hline
        $m$ & $\mathbb{R}$ & Mass \\
        $\mJ$ & $\mathbb{R}^{3 \times 3}$ & Inertia \\
        $\vp$ & $\mathbb{R}^3$ & Position \\
        $\mathbf{q}$ & $\mathbb{R}^4$ & Attitude unit quaternion \\
        $\vv$ & $\mathbb{R}^3$ & Global velocity \\
        $\vomega$ & $\mathbb{R}^3$ & Body angular velocity \\\hline
        $\Omega_i$ & $\mathbb{R}_{\geq 0}$ & $i$-th squared motor speed \\
        $c_{f_i}$ & $\mathbb{R}$ & $i$-th propeller's shape force parameter \\
        $c_{\tau_i}$ & $\mathbb{R}$ & $i$-th propeller's shape torque parameter \\
        $\mathbf{z}_{\Omega_i}$ & $\mathbb{R}^3$ & Unit vector parallel to the $i$-th rotor's rotation axis \\
        $\mathbf{p}_{\Omega_i}$ & $\mathbb{R}^3$ & $i$-th rotor's position in the body frame \\
        $\mathbf{F}$ & $\mathbb{R}^{3 \times n}$ & Force allocation matrix \\
        $\mathbf{M}$ & $\mathbb{R}^{3 \times n}$ & Torque allocation matrix \\
        $\emph{g}$ & $\mathbb{R}$ & Acceleration due to gravity, $\mathbf{g} = (0,0,-\emph{g})^\top$ \\
        $\mathbf{R}$ & $SO(3)$ & Body-to-global rotation matrix \\
        $\mathbf{f}_a$ & $\mathbb{R}^3$ & External forces acting on the \ac{UAV} \\
        $\bm{\tau}_a$ & $\mathbb{R}^3$ & External torques acting on the \ac{UAV} \\\hline
        $\mathbf{f}_{s_i}$ & $\mathbb{R}^3$ & External force from aerodynamic surface $s_i$  \\
        $\mathbf{l}_{s_i}$ & $\mathbb{R}^3$ & Displacement from \acs{CoM} to center of lift for $s_i$ \\
        $\rho$ & $\mathbb{R}$ & Air density \\
        $\mathbf{v}_{s_i}$ & $\mathbb{R}^3$ & Relative wind at aerodynamic center in $s_i$ frame \\
        $S_i$ & $\mathbb{R}$ & Surface area of the $i$-th surface \\
        $c_{L_i}$ & $\mathbb{R}$ & Lift coefficient for the $i$-th surface \\
        $c_{D_i}$ & $\mathbb{R}$ & Drag coefficient for the $i$-th surface \\
        $\mathbf{e}_{D_i}$ & $\mathbb{R}^3$ & Unit vector in $i$-th surface drag direc. $(\mathbf{v}_{s_i}/\lVert \mathbf{v}_{s_i} \rVert)$ \\ 
        $\mathbf{e}_{L_i}$ & $\mathbb{R}^3$ & Unit vector in $i$-th surface lift direction ($\perp$ to $\mathbf{e}_{D_i}$) \\
        $\mathbf{R}_{s_i}$ & $SO(3)$ & Body-to-surface rotation matrix for $s_i$ \\\hline
        $\mathbf{f}_e$ & $\mathbb{R}^3$ & Body frame environment force on end-effector \\
        $\bm{\tau}_e$ & $\mathbb{R}^3$ & Body frame environment torque on end-effector \\
        $\mathbf{p}_e$ & $\mathbb{R}^3$ & End-effector tip's position in body frame \\\hline
    \end{tabular}
    \end{center}
}


\subsection{Fixed-wings}
\label{sec:fixedWing}

\textbf{Basic}: 
For fixed-wing~\acp{UAV}, similarly to multirotors, equations of motion derive from Newton-Euler formalism. We extend (\ref{eq:multirotorDynamics}) with the following external aerodynamic forces and torques.

\vspace*{0.6mm}
\quad\textit{External Aerodynamic Forces:}
$\mathbf{f}_a = \sum_i^n \mathbf{f}_{s_i}$, 
\vspace*{0.6mm}

\quad\textit{External Aerodynamic Torques:}
$\bm{\tau}_a = \sum_{i}^n (\mathbf{l}_{s_i} \times \mathbf{f}_{s_i})$,
\vspace*{0.6mm}

\quad\textit{Surface Forces:}
$\mathbf{f}_{s_i}= \frac{1}{2} \rho \lvert \mathbf{v}_{s_i} \rvert^2 S_i \left(c_{L_i} \mathbf{e}_{L_i} + c_{D_i} \mathbf{e}_{D_i} \right)$,
\vspace*{0.6mm}

\quad\textit{Relative Wind:}
$\mathbf{v}_{s_i} = -\mathbf{R}_{s_i}^\top \left( \mathbf{R}(\mathbf{q})^\top \mathbf{v} + \bm{\omega} \times \mathbf{l}_{s_i} \right)$.
\vspace*{0.6mm}

For control surfaces with a single \ac{DoF}, both $\mathbf{R}_{s_i}$ and $\mathbf{l}_{s_i}$ depend on the control surface deflection, $\delta_i$, which is treated as an additional control variable, where $\mathbf{u}_{\delta}=(\delta_1,\hdots,\delta_c)^\top$.
Typical aerodynamic surfaces include wings and vertical or horizontal stabilizers; typical control surfaces are ailerons, rudder, and elevator \cite{basescu2020direct, sobolic2009nonlinear}. 

\textbf{Lift and Drag}: For fixed-wings, lift and drag forces play pivotal roles in vehicle simulation. The coefficients $c_{L_i}$ and $c_{D_i}$ can be derived by combining 
airfoil data at low angle-of-attack with high angle-of-attack approximations obtained from flat-plate theory \cite{hoerner1985fluid}. Additionally, these coefficients can also be partly or entirely data-driven~\cite{basescu2023precision}. 

\textbf{Wind}: Wind is modeled as a spatio-temporal-varying additive velocity contributing directly to $\mathbf{v}_{s_i}$ \cite{basescu2022onboard}.

\textbf{Interactions}: Ground effects significantly impact fixed-wing \acp{UAV} during landing maneuvers.
Data-driven approaches effectively capture this effect by relating $c_{L_i}$ to ground proximity \cite{ambati2017robust}.



\subsection{Aerial manipulators}
\label{sec:aerialManipulation}

\textbf{Basic}: 
Aerial manipulators consist of an aerial base and an end-effector 
(e.g. rigid tools, articulated arms, cables). Here, we include a general description, 
excluding advanced techniques like soft manipulation \cite{gomez2019current}.

\textbf{Rigid Tools}: Aerial manipulators with rigid tools follow (\ref{eq:multirotorDynamics}) with adjustments for the tool's inertial properties. The terms $\mathbf{f}_a$ and $\bm{\tau}_a$ account for the wrench applied at the end-effector. 

\vspace*{0.6mm}
\quad\textit{External Force:} $\mathbf{f}_a = \mathbf{R}(\mathbf{q}) \mathbf{f}_e$, 
\vspace*{0.6mm}

\quad\textit{External Torque:} $\bm{\tau}_a = \mathbf{p}_e \times \, \mathbf{f}_e + \bm{\tau}_e$.
\vspace*{0.6mm}

\textbf{Articulated Arms}: For manipulators with articulated arms, dynamics become more complex \cite{meng2020survey}. 

\textbf{Cables}: Cables are typically modeled as massless rigid or elastic elements attached via passive spherical joints to the robot's \ac{CoM} \cite{gabellieri2023equilibria}. In this scenario,  $\mathbf{f}_a = \mathbf{R}(\mathbf{q}) \mathbf{f}_e$ accounts for the cable's force.  
In this formulation, external torque applied to the robot by the cable is excluded ($\bm{\tau}_a = \bm{0}_{3\times1}$) \cite{meng2020survey}.

\end{tcolorbox}


\vspace*{-1mm}
\section{UAV Simulators} 
\label{sec:descriptionAerialSimulators}

\begin{table}[tb]
    \centering
    \vspace{-2mm}
    \caption{Selection Criteria for \ac{UAV} Simulators}
    \vspace*{-5mm}
    \label{tab:sim_selection}
    \begin{center}
    \renewcommand{\arraystretch}{1.3}
    \begin{tabular}{c | p{5.75cm}}
        \hline
        \textbf{Criteria} & \textbf{Decision Factors} \\
        \hline \hline
        Physics Fidelity & Required fidelity of physics and dynamics model for the intended use case \\ \hline
        Visual Fidelity & Level of realism in images (e.g., for computer vision or \ac{ML} applications) \\ \hline
        Autopilots & Compatibility with common autopilots like PX4 and ArduPilot, useful for~\ac{SITL} and~\ac{HITL} testing \\ \hline
        Multiple Vehicles & Capability to concurrently simulate vehicles \\ \hline 
        Heterogeneity & Integration possibilities with other platforms \\ \hline
        Sensors & Integration support for common sensors (e.g., cameras, IMU, GPS, LiDAR, optical flow) 
        \\ \hline 
        \ac{UAV} Models & Support of common~\ac{UAV} models and ease of integrating new models \\ \hline 
        Simulation Speed & Real-time speed and ability to run in super real-time, crucial for learning applications \\ \hline 
        \acsp{API} & Compatibility with programming languages, middleware like~\acs{ROS}, and packages such as OpenAI Gym (now Gymnasium) \\ \hline
        Integration & Ease of getting started and development, license type, and maintenance status of the software \\ \hline 
    \end{tabular}
    \end{center}
    \vspace*{-8.5mm}
\end{table}

A primary consideration when selecting a simulator 
is the specific application domain
and whether the available simulators offer the necessary features and sensors tailored to that domain.
Additionally, compatibility with common autopilots (e.g. PX4 and ArduPilot) or standardized research hardware (e.g. the Crazyflie, a palm sized \ac{UAV}) is often a consideration to enable rapid simulation to hardware transfer. Many specialized simulators have emerged tailored to these components.
Drawing from our own experiences and the referenced literature, we compiled a set of selection criteria and decision factors that are regularly considered when evaluating \ac{UAV} simulators. These comparative points are outlined in Table~\ref{tab:sim_selection}. 
Table~\ref{tab:sim_categories} categorizes a range of \ac{UAV} simulators based on their key elements. 
We consider the following categories: a ``Universal Simulator'' is targeted toward simulating (potentially simultaneously) a wide variety of platforms (e.g. ground and aerial vehicles, manipulators, etc.); ``Sensor-Focused'' targets simulating realistic performance of a particular sensor (e.g. for photorealism or accurate LiDAR measurements); ``Learning-Focused'' targets compatibility with \ac{ML} architectures (e.g. for training a new behavior or learning a dynamics model); ``Dynamics-Focsued'' emphasizes thorough dynamics models for a particular class of systems (e.g. aerial manipulation, like RotorTM and HIL-airmanip, or fixed-wings, like PyFly); ``Swarming'' simulates a group of vehicles; ``Part of Flight Stacks'' are simulators with flight stack integration; and ``Flight Simulators'' are flight emulators often for large-scale human-piloted aircraft. Flight simulators are less often used for robotics applications, however some have been adapted for this purpose due to their realistic flight experience and we have included them here for completeness.
Some simulators may belong to multiple categories.

What follows offers a concise overview of each simulator featured in Table~\ref{tab:sim_categories}. 
We present each simulator in a subsection corresponding to one of its major categories, with references in the other categories corresponding to Table~\ref{tab:sim_categories}.

\begin{table}[tb]
    \centering
    \caption{\ac{UAV} Simulators Categorization}
    \vspace*{-5mm}
    \label{tab:sim_categories}
    \begin{center}
    \renewcommand{\arraystretch}{1.3}
    \renewcommand{\tabcolsep}{1.35mm}
    \begin{tabular}{c c | p{6.28cm}}
        \hline
        \multicolumn{2}{c|}{\textbf{Category}} & \textbf{Simulators} \\ \hline \hline
        \multicolumn{2}{C{2.0cm}|}{\multirow{2.5}{*}{\makecell{Universal\\Simulators}}} & Gazebo Classic (RotorS, CrazyS), Gazebo (PX4 SITL, ArduPilot SITL), Isaac Sim/Gym (Pegasus, Aerial Gym), Webots, CopelliaSim, MuJoCo \\ \hline
        \multirow{2}{3.1em}{Sensor-Focused} & Vision &  AirSim, Flightmare, FlightGoggles, FastSim \\ \cline{2-3}
        ~ & LiDAR & MARSIM \\ \hline
        \multicolumn{2}{C{2.0cm}|}{\multirow{2.5}{*}{Learning-Focused}} & Isaac Gym (Aerial Gym), MuJoCo, PRL4AirSim, Flightmare, gym-pybullet-drones, safe-control-gym, QuadSwarm, fixed-wing-gym, QPlane, PyFlyt \\ \hline
        \multicolumn{2}{C{2.0cm}|}{Dynamics-Focused} & RotorTM, MATLAB UAV Toolbox, PyFly, ARCAD, HIL-airmanip, RotorPy, RflySim, Agilicious \\ \hline
        \multicolumn{2}{C{2.0cm}|}{Swarming} & gym-pybullet-drones, QuadSwarm, Potato \\ \hline
        \multicolumn{2}{C{2.0cm}|}{Part of Flight Stacks} & Agilicious, MRS UAV System, CrazyChoir, Crazyswarm2, Aerostack2, CrazySim, sim\_cf2 \\ \hline
        \multicolumn{2}{C{2.0cm}|}{Flight Simulators} & X-Plane (X-PlaneROS), QPlane, FlightGear, RealFlight \\ \hline
    \end{tabular}
    \end{center}
    \vspace*{-8.5mm}
\end{table}




\subsection{Universal simulators}
\label{sec:universal_sims}

\textit{Gazebo Classic} \cite{KoenigHoward2004Design} is an open-source, continuously maintained, versatile research simulation platform with a modular design, accommodating different physics engines, sensors, and 3D world creation. 
Particularly noteworthy is its suitability for aerial manipulator tasks, owing to its ease of creating contact surfaces with customizable frictions \cite{suarez2020compliant, DimmigKobilarov2023Small}. 
\textit{RotorS} \cite{FurrerSiegwart2016RotorS}, built on top of Gazebo Classic, offers a modular framework for designing \acp{UAV} and developing control algorithms, particularly focusing on simulating the vehicle dynamics.
\textit{CrazyS} \cite{SilanoIannelli2018CrazyS}, an extension of RotorS, focuses on modeling the Crazyflie 2.0 quadrotor. However, both RotorS and CrazyS have limited perception-related capabilities. 
The new \textit{Gazebo} \cite{Gazebo}, formerly known as Ignition, is the successor of Gazebo Classic and incorporates quadrotor dynamics and control inspired by the RotorS project. 
Gazebo enables dynamic loading and unloading of environment components, addressing challenges faced by Gazebo Classic in replicating large, realistic environments. 
Moreover, Gazebo offers improved interfaces for simulating radio communication between multiple \acp{UAV}.
\textit{PX4 \ac{SITL} Gazebo} \cite{PX4SITL} and \textit{ArduPilot SITL Gazebo} \cite{ArduPilotSITL} are built on top of Gazebo and include the latest support for \ac{SITL} testing with PX4 and ArduPilot, respectively. These simulators do not depend on \acs{ROS} and support simulating a large number of vehicles and sensors. 

\textit{Isaac Sim} \cite{IsaacSim}, developed by NVIDIA, is a photorealistic high-fidelity simulator for a variety of platforms. 
\textit{Pegasus Simulator} \cite{jacinto2024pegasus} is an open-source extension to Isaac Sim that includes an extended multirotor dynamics model, simulating multiple vehicles in parallel, integration with PX4 and \acs{ROS} 2, and additional sensors (magnetometer, GPS, and barometer).
\textit{Isaac Gym} \cite{makoviychuk2021isaac} is NVIDIA's library for GPU-accelerated \ac{RL} simulations and uses more basic rendering than Isaac Sim. 
\textit{Aerial Gym} \cite{KulkarniAlexis2023Aerial} is an open-source extension to Isaac Gym Preview Release 4 notable for its capability to parallelize the simulation of thousands of multirotors and includes customizable obstacle randomization.

\textit{Webots} \cite{MichelMichel2004Cyberbotics} is an open-source, versatile robotics simulator known for its wide range of robotic platforms. While Webots primarily focuses on ground-bound robots, it also features two quadrotor models with simplified aerodynamic physics. Webots uses \acs{ODE} for physics simulation, refer to \cite{Erez2015ICRA} for a comprehensive analysis. 
Notably, Webots has been used for 
innovative vehicle designs, such as a triphibious robot in \cite{GuXia2021Design}.

\textit{CoppeliaSim} \cite{RohmerFreese2013CoppeliaSim}, previously known as V-REP, is a versatile robotics simulator with support for a wide range of programming languages and physics engines. 
Selecting the appropriate physics engine is crucial to avoid undesirable outcomes, such as velocity or position jumps, unrealistic collision behaviors, and erratic sensor outputs \cite{Erez2015ICRA}. 
CoppeliaSim has been used for applications such as 
\ac{UAV} obstacle avoidance \cite{UdvardyBotos2020Simulation}.

\textit{MuJoCo} \cite{TodorovTassa2012MuJoCo}, see Sec.~\ref{sec:learning_focused_sims}.



\subsection{Sensor-focused simulators}
\label{sec:sensor_focused_sims}

\textit{AirSim} \cite{ShahKapoor2018AirSim} is a Microsoft-led project built on the Unreal Engine, 
offering various sensors, a weather API, and compatibility with open-source controllers (e.g., PX4 and ArduPilot). AirSim primarily serves as a platform for AI research, providing platform-independent \acsp{API} for data retrieval and vehicle control. Notably, AirSim demands substantial computing power compared to other simulators. 
\textit{PRL4AirSim} \cite{SaundersLil2023Parallel} is an extension for efficient parallel training in \ac{RL} applications. 
The original AirSim is open-source, but will no longer be supported by Microsoft. Their focus has shifted to \textit{Project AirSim}, which will be released under a commercial license. 

\textit{Flightmare} \cite{SongScaramuzza2021Flightmare} is a versatile simulator with two main components: a Unity-based rendering engine and a physics model, both designed for flexibility and independent operation. The rendering engine can generate realistic visual information 
and simulate sensor noise, environmental dynamics, and lens distortions with minimal computational overhead. Similarly, the physics model allows users to control robot dynamics, ranging from basic noise-free \acp{UAV} models to advanced rigid-body dynamics with friction and rotor drag, or even real platform dynamics. 
Flightmare is extensively used for \ac{ML} applications, such as for autonomous drone racing \cite{SongScaramuzza2023Reaching}. 

\textit{FlightGoggles} \cite{GuerraKaraman2019FlightGoggles}, similarly to Flightmare, is an open-source simulator focused on photorealistic simulation. FlightGoggles combines two key elements: (i) photogrammetry for realistic simulation of camera sensors, 
and (ii) virtual reality to integrate real vehicle motion and human behavior in the simulations. 
FlightGoggles is built around the Unity engine and includes multirotor physics with motor dynamics, basic vehicle aerodynamics, and IMU bias dynamics. 
A key feature of FlightGoggles is the ``vehicle-in-the-loop simulation,'' 
where the vehicle is flown in a motion capture system, camera images and exteroceptive sensors are simulated in Unity, and collision detection is based on the real-world vehicle's pose. 

\textit{FastSim} \cite{CuiXu2024FastSim}  is a high-fidelity,  modular simulation framework built on Unity (for photorealistic simulation). Configurable modules include sensors, algorithms, robots, worlds, and display tools. Implemented sensors include IMU, RGB, depth, segmentation, and event cameras. Both \ac{SITL} and \ac{HITL} are supported.  Demonstrated applications include vision-based localization, motion planning, swarming, and \ac{ML}.

\textit{MARSIM} \cite{KongZhang2023MARSIM} is an open-source C/C++ library primarily focused on accurately simulating LiDAR measurements for \acp{UAV}. It constructs depth images from point cloud maps and interpolates them to obtain LiDAR point measurements. The simulator is designed for lightweight computation and offers access to 10 high-resolution environments, including forest, historic building, office, parking garage, and indoor settings. 



\subsection{Learning-focused simulators}
\label{sec:learning_focused_sims}

\textit{MuJoCo} \cite{TodorovTassa2012MuJoCo}, or Multi-Joint dynamics with Contact, is a frequently employed physics engine and simulator in \ac{ML} applications. It offers interactive visualization rendered with OpenGL and encompasses various platforms, including a \ac{UAV} model (i.e. the Skydio X2 quadrotor). 

\textit{safe-control-gym} \cite{YuanSchoellig2022Safe} is an open-source safety-focused \ac{RL} environment and benchmark suite, built using the PyBullet physics engine \cite{CoumansBai2016PyBullet}, for comparing control and \ac{RL} approaches. Three dynamics systems (cart-pole, 1D and 2D quadrotor) and two control tasks (stabilization and trajectory tracking) are included. 
This simulation environment supports model-based and data-based approaches, expresses safety constraints, and captures real-world properties (such as uncertainty in physical properties and state estimation). 

\textit{PyFlyt} \cite{TaiPhang2023PyFlyt} is an open-soruce simulator using the PyBullet physics engine \cite{CoumansBai2016PyBullet}. 
For \ac{RL} research, PyFlyt includes compatibility with Gymnasium \cite{towers_gymnasium_2023} and PettingZoo \cite{TerryRavi2021PettingZoo} (for multi-agent \ac{RL}). Basic \ac{UAV} components are expressed modularly to allow for flexible construction of \acp{UAV}, including support for multirotors and fixed-wings. 

\textit{Isaac Gym (Aerial Gym)} \cite{makoviychuk2021isaac} (\hspace{1sp}\cite{KulkarniAlexis2023Aerial}), see Sec.~\ref{sec:universal_sims}. 

\textit{PRL4AirSim} \cite{SaundersLil2023Parallel}, see Sec.~\ref{sec:sensor_focused_sims}.

\textit{Flightmare} \cite{SongScaramuzza2021Flightmare}, see Sec.~\ref{sec:sensor_focused_sims}.

\textit{gym-pybullet-drones} \cite{PaneratiSchoellig2021Learning}, see Sec.~\ref{sec:swarming_sims}.

\textit{QuadSwarm} \cite{HuangSukhatme2023QuadSwarm}, see Sec.~\ref{sec:swarming_sims}.

\textit{fixed-wing-gym} \cite{BoehnJohansen2019Deep}, see Sec.~\ref{sec:dynamics-focused_sims}.

\textit{QPlane} \cite{RichterCalix2021QPlane}, see Sec.~\ref{sec:flight_sims}. 



\subsection{Dynamics-focused simulators}
\label{sec:dynamics-focused_sims}

\textit{RotorTM} \cite{LiLoianno2023RotorTM} is an open-source simulator for aerial object manipulation. 
Notably, this simulator considers cable-suspended loads and passive connection mechanisms between multiple vehicles, a feature lacking in other common simulators. In RotorTM, the cables are considered massless and connected to the robot's \ac{CoM}. They can transition from taut to slack during task execution, allowing users to customize the number of robots and the type of payload (e.g. rigid body or point mass). Additionally, the simulator accommodates scenarios where aerial robots are rigidly attached to the load. RotorTM assumes negligible drag on the payload and aerial robot and disregards aerodynamic effects, 
considering rotor dynamics to be significantly faster than other factors.

\textit{MATLAB \ac{UAV} Toolbox} \cite{MATLABMATLABUAV} is a general purpose toolbox for designing, simulating, testing, and deploying \acp{UAV} within MATLAB. It includes tools for algorithm development, flight log analysis, and simulation. The simulation capabilities include a cuboid simulation for quickly constructing new scenarios and a photorealistic 3D simulation environment with synthesized camera and LiDAR readings. The toolbox includes an interface for deploying directly to hardware through PX4-based autopilots. Additionally, the MAVLink protocol is supported. Researchers have explored using the MATLAB \ac{UAV} Toolbox with flight simulators \cite{HorriPietraszko2022Tutorial} such as X-Plane, FlightGear, and, in other works (as mentioned in \cite{HorriPietraszko2022Tutorial}) RealFlight. 

\textit{PyFly} \cite{BoehnJohansen2019Deep} is an open-source Python simulator designed for fixed-wing aircraft. It includes a 6-\ac{DoF} aerodynamic model, wind effects, and stochastic turbulence. \textit{fixed-wing-gym} \cite{BoehnJohansen2019Deep} is an OpenAI Gym wrapper specifically tailored for PyFly, aiming at facilitating \ac{RL} applications. 

\textit{ARCAD} \cite{KeipourScherer2023UAS}, or AirLab Rapid Controller and Aircraft Design, is an open-source MATLAB simulator for fully-actuated multirotors. Its primary goals are to expedite the modeling, design, and analysis of new aircraft and controllers and the visualization of tasks involving physical interactions, including controlled force-based tasks like writing text on a wall.

\textit{HIL-airmanip} \cite{cuniato2021hardware} offers a distinctive environment for simulating physical interactions between humans and aerial robots, enabling real-time human involvement. In this simulator, the forces exchanged between the human operator and a haptic interface are accurately measured and then transmitted to an aerial manipulator, which is modeled within the RotorS environment. This robotic system consists of a quadrotor combined with a 6-\ac{DoF} arm mounted beneath it. 

\textit{RotorPy} \cite{FolkKumar2023RotorPy} is an open-source Python simulator meant to be lightweight and focused on providing a comprehensive quadrotor model. Its development emphasizes accessibility, transparency, and educational value, initially created as a teaching tool for a robotics course at the University of Pennsylvania. In \cite{FolkKumar2023RotorPy}, the simulator's 
quadrotor model is extensively detailed, including 6-\ac{DoF} dynamics, aerodynamic wrenches, actuator dynamics, sensors, and wind models. The model's validity is verified using a Crazyflie performing agile maneuvers. Additionally, the simulator includes a Gymnasium environment for \ac{RL} applications.

\textit{RflySim} \cite{DaiCai2021RFlySim} is a model-based design toolchain in MATLAB that can be used for \ac{SITL} and \ac{HITL} testing with an emphasis on model credibility and extensibility. Unreal is used for visual scene rendering, PX4 compatibility is built in, and multiple \ac{UAV} models are supported.

\textit{Agilicious} \cite{FoehnScaramuzza2022Agilicious}, see Sec.~\ref{sec:flight_stack_sims}.



\subsection{Swarming simulators}
\label{sec:swarming_sims}

\textit{gym-pybullet-drones} \cite{PaneratiSchoellig2021Learning} is an open-source environment designed for simulating multiple quadrotors with PyBullet \cite{CoumansBai2016PyBullet} physics, tailored for research that combines control theory and \ac{ML}. 
This library has interfaces for multi-agent and vision-based \ac{RL} applications, utilizing the Gymnasium \acsp{API} \cite{towers_gymnasium_2023}. It supports the definition of various learning tasks on a Crazyflie platform. Notably, gym-pybullet-drones includes realistic collisions and aerodynamic effects (e.g. drag, ground effect, and downwash). 
It includes example \ac{RL} workflows for single agent and multi-agent scenarios, leveraging Stable-baselines3 \cite{RaffinDormann2021Stable}.

\textit{QuadSwarm} \cite{HuangSukhatme2023QuadSwarm} is an open-source Python library for multi-quadrotor simulation in \ac{RL} applications, emphasizing fast simulation and the transfer of policies from simulation to the real-world. QuadSwarm provides diverse training scenarios and domain randomization to support \ac{RL} applications, showcasing zero-shot transfer of \ac{RL} control policies for single and multi-quadrotor scenarios. The physics model is based on the Crazyflie platform, with OpenGL used for rendering.

\textit{Potato} \cite{LiRen2023Potato} is based on data-oriented programming 
for large-scale swarm simulations. Like Isaac Gym, Potato relies on GPU computation rather than CPU. 
It includes basic dynamics for fixed-wing drones, quadrotors, and cars. Potato is not currently open-source, but the authors expressed the intention to open-source the quadrotor piece in the future. 



\subsection{Simulators part of flight stacks}
\label{sec:flight_stack_sims}

\textit{Agilicious} \cite{FoehnScaramuzza2022Agilicious} contains a hardware description for a quadrotor with a Jetson TX2 and a software library specifically meant for autonomous and agile quadrotor flight. For simulation, it has a custom modular simulator that incorporates highly accurate aerodynamics based on blade-element momentum theory or with other tools like RotorS, \ac{HITL} setups, and rendering engines such as Flightmare.
The stack uses a custom license, but is free to use for academics after registration.

\textit{MRS UAV System} \cite{BacaSaska2021MRS} is a flight stack designed for replicable research through realistic simulations and real-world experiments. 
Its software stack includes a simulation environment built on Gazebo Classic, CoppeliaSim, or their MRS-multirotor-simulator for quadrotor dynamics, complete with realistic sensors and models. A key feature is its 
ongoing active use and maintenance. Moreover, this stack is frequently used for teams of multirotors. 

\textit{CrazyChoir} \cite{PichierriNotarstefano2023CrazyChoir} is an open-source modular \acs{ROS} 2 framework designed for conducting realistic simulations and experiments involving cooperating Crazyflie drones.
For simulation, it builds on Webots with a \ac{SITL} of the Crazyflie firmware.

\textit{Crazyswarm2} \cite{PreissAyanian2017Crazyswarm} is an open-source framework designed for controlling large indoor quadrotor swarms, specifically utilizing Crazyflie drones, similarly to CrazyChoir. For simulation it also relies on \ac{SITL} of the firmware with a modular simulation framework that currently supports pure visualization or an \textit{ad hoc} Python-based physics simulation. 

\textit{Aerostack2} \cite{FernandezCortizasCampoy2023Aerostack2} is a versatile open-source flight stack designed to be compatible with various \ac{UAV} platforms, including PX4, ArduPilot, DJI, and Crazyflie.
For simulation purposes, Aerostack2 utilizes Gazebo with custom sensors. However, it does not currently have support for (S/H)ITL simulations.

\textit{CrazySim} \cite{LlanesCoogan2024CrazySim} is a \ac{SITL} simulation pipeline for swarming research with the Crazyflie quadrotor. An individual instance of the Crazyflie flight stack is run for each \ac{UAV}, with sensors and communication simulated in Gazebo, in order to test firmware code in simulation. Additionally, radio communication between the \acp{UAV} and ground station, including delays, are simulated. \textit{sim\_cf2} \cite{simcf2} is very similar to CrazySim for \ac{SITL} simulation, but instead uses Gazebo Classic. Additionally, sim\_cf2 is more integrated with \acs{ROS}~2 versus CrazySim offers a Gazebo plugin without \acs{ROS}~2 dependencies.



\subsection{Flight simulators}
\label{sec:flight_sims}

\textit{X-Plane} \cite{XPlane} is a commercial cross-platform flight simulator. As with most flight simulators, the primary audience is pilots. 
X-Plane emphasizes realistic dynamics (focused on large-scale fixed-wing and helicopter vehicles) and includes simulated weather, wind, and lighting conditions. 
\textit{X-PlaneROS} \cite{navarro2024sorts} is a X-Plane \acs{ROS} 1 wrapper for controlling large-scale fixed-wing vehicles, extracting aircraft data from the simulator, and enables human-robot interaction. 
\textit{QPlane} \cite{RichterCalix2021QPlane} is a \ac{RL} toolkit for fixed-wing simulation that can use external flight simulators, such as X-Plane and/or FlightGear. 

\textit{FlightGear} \cite{PerryPerry2004flightgear} is an open-source, user supported, cross-platform flight simulator. 
Some researchers have explored using this flight simulator for \ac{UAV} simulation, such as in \cite{PrabowoTriputra2015Hardware}. 

\textit{RealFlight} \cite{RealFlight} is a commercial Windows \ac{RC} flight simulator 
that includes small multirotor and fixed-wing vehicles.
Researchers have adopted this flight simulator for \ac{UAV} simulation, such as in \cite{CarlsonPapachristos2021MiniHawk}.



\section{UAV Simulator Comparison}
\label{sec:comparisonAerialSimulators}

In Table~\ref{tab:pro_con}, we summarize prominent positive and negative decision factors for all included simulators.
We then provide three more detailed tables, for a subset of the simulators, that compare essential features of aerial simulators, using the selection criteria discussed in Table~\ref{tab:sim_selection}. 
We exclude simulation speed as this varies depending on the environment's complexity, physics model, rendering quality, and computer hardware. We encourage the reader to preform this test (as needed) for their application space.
The detailed tables include extensively utilized simulators for aerial vehicles. 
For brevity, we omitted simulators that are less versatile or relatively new, leading to limited adoption. 

\begin{figure}[tb]
    \begin{center}
	\begin{subfigure}[t]{0.40\columnwidth}
    	\centering
    	\includegraphics[width=\columnwidth]{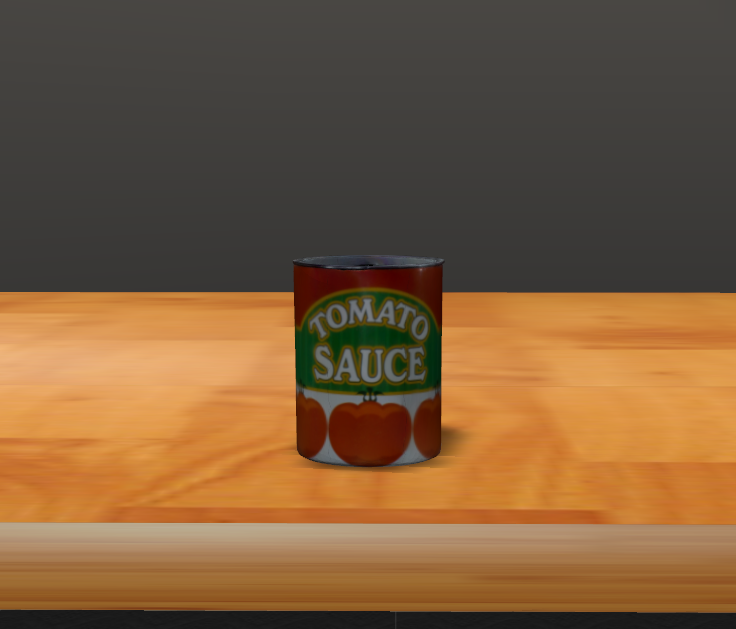}
    	\captionsetup{font=scriptsize}
            \vspace*{-1.0em}
            \caption{Gazebo Classic}
    	\label{fig:gazebo_camera}
	\end{subfigure}
    \hspace*{2mm}
	\begin{subfigure}[t]{0.40\columnwidth}
    	\centering
    	\includegraphics[width=\columnwidth]{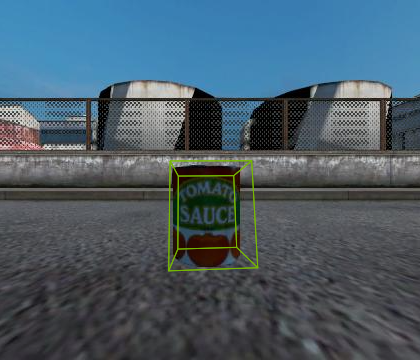}
    	\captionsetup{font=scriptsize}
            \vspace*{-1.0em}
            \caption{Flightmare}
    	\label{fig:flightmare_camera}
	\end{subfigure}
    \caption{Toy can in Gazebo Classic and Flightmare simulation environments. In (b), the green bounding box denotes a detection by~\acf{DOPE}~\cite{TremblayBirchfield2018Deep}. 
    The comparable image in the Gazebo Classic simulation did not yield detections by~\ac{DOPE} due to its lower visual fidelity. Gazebo Classic utilizes OGRE for rendering, while Flightmare uses Unity.}
    \label{fig:camera_views}
    \end{center}
    \vspace*{-0.8em}
\end{figure} 

Table~\ref{tab:sim_features} presents a comparison of notable features within the simulation environments. 
Different applications may require different physics fidelity, such as for environmental contact or physical accuracy. There is often a trade-off between speed and accuracy. We include the physics engines used in each simulator in Table~\ref{tab:sim_features} and refer the reader to \cite{Erez2015ICRA} as needed for a comprehensive analysis of different physics engines.
We denote the rendering capabilities of the simulators. Notably, OpenGL and OGRE rendering are often regarded as having low visual fidelity, while Vulkan, Unity, and Unreal rendering are considered to offer high visual fidelity. 
Figure~\ref{fig:camera_views} depicts a comparison of camera views in an OGRE environment versus a Unity environment.
The \ac{OS} subcategories are Linux, Windows, and Mac, denoted as ``L,'' ``W,'' ``M,'' respectively. 
``\ac{RL}'' is included as an interface to indicate specialization for \ac{RL} applications. The category denoted as (S/H)ITL includes interfaces for both \ac{SITL} and \ac{HITL} capabilities, with ``CF'' representing Crazyflie.
Finally, we denote the license and whether each simulator is open-source and refer the reader to \cite{Ballhausen2019Free} for more information on various license types.
Lastly, we include a column ``Active'' 
\FloatBarrier
\begin{table*}[tbh]
    \centering
    \caption{Prominent Decision Factors to Consider for each Simulator}
    \vspace*{-4.5mm}
    \label{tab:pro_con}
    \begin{center}
    \renewcommand{\arraystretch}{1.3}
    \begin{tabular}{c | L{7.0cm} | L{5.5cm} | c}
         \hline
         \textbf{Simulator}&  \textbf{Positive Factors}& \textbf{Negative Factors} & \textbf{Ref.}\\\hline\hline
         Gazebo Classic & Built into \acs{ROS} 1, large community, diverse platform support & Low visual fidelity, slow comp., phasing out & \cite{KoenigHoward2004Design} \\\hline
         RotorS (with Gazebo Classic) & Key \ac{UAV} modeling and algorithms for Gazebo Classic & Built on Gazebo Classic (default in Gazebo) & \cite{FurrerSiegwart2016RotorS} \\\hline
         CrazyS (with Gazebo Classic) & Crazyflie specific extension for Gazebo Classic & Built on Gazebo Classic, limited-perception & \cite{SilanoIannelli2018CrazyS} \\\hline
         Gazebo & Built into \acs{ROS} 2, dynamic environment loading & Low visual fidelity & \cite{Gazebo} \\\hline
         PX4 \ac{SITL} Gazebo & PX4 \ac{SITL}, large community, extensive support & PX4 exclusive & \cite{PX4SITL} \\\hline
         ArduPilot \ac{SITL} Gazebo & ArduPilot \ac{SITL}, large community, extensive support & ArduPilot exclusive & \cite{ArduPilotSITL} \\\hline
         Isaac Sim & Photorealistic, high-fidelity, NVIDIA supported  & Not open-source & \cite{IsaacSim} \\\hline
         Pegasus (with Isaac Sim) & Extended multirotor dynamics and sensors, PX4 integration & Back end (Isaac Sim) is not open-source & \cite{jacinto2024pegasus} \\\hline
         Isaac Gym &  GPU-accelerated \ac{RL} simulations, NVIDIA supported & Not open-source & \cite{makoviychuk2021isaac} \\\hline
         Aerial Gym (with Isaac Gym) & Highly parallelizable, GPU-enabled geometric controllers  & Back end (Isaac Gym) is not open-source & \cite{KulkarniAlexis2023Aerial} \\\hline
         Webots & User-friendly, diverse platform support & Focused on ground robots, basic aerodynamics & \cite{MichelMichel2004Cyberbotics} \\\hline
         CoppeliaSim & Large number of physics engines and platforms & Not fully open-source & \cite{RohmerFreese2013CoppeliaSim} \\\hline
         AirSim & Photorealistic, large sensor and weather support & Not actively maintained, substantial compute & \cite{ShahKapoor2018AirSim} \\\hline
         PRL4AirSim (with AirSim) & Parallel training for \ac{RL} using AirSim & Built on AirSim & \cite{SaundersLil2023Parallel} \\\hline
         Flightmare & Photorealistic, computation minimized, built for \ac{ML} & Not actively maintained & \cite{SongScaramuzza2021Flightmare} \\\hline
         FlightGoggles & Photorealistic, vehicle-in-the-loop simulation & Not actively maintained & \cite{GuerraKaraman2019FlightGoggles} \\\hline
         FastSim & Photorealistic, extensive sensors, modular, generalized & Brand-new, code not released at time of writing & \cite{CuiXu2024FastSim} \\\hline
         MARSIM & LiDAR measurement simulation, lightweight computation & Solely LiDAR focused & \cite{KongZhang2023MARSIM} \\\hline
         MuJoCo & Common for ML applications, parallel processing & Low visual fidelity & \cite{TodorovTassa2012MuJoCo} \\\hline
         safe-control-gym & Safety-focused \ac{RL}, benchmarks, control vs \ac{RL} comparisons & Simplistic dynamical systems & \cite{YuanSchoellig2022Safe} \\\hline
         PyFlyt & Gymnasium \& PettingZoo integration, modular \acp{UAV} & Low visual fidelity & \cite{TaiPhang2023PyFlyt} \\\hline
         RotorTM & Cable-suspended loads and passive vehicle connections & Cables massless, negligible drag on payload & \cite{LiLoianno2023RotorTM} \\\hline
         MATLAB \ac{UAV} Toolbox & General toolbox for \ac{UAV} design to deployment & Not open-source, \ac{UAV} exclusive & \cite{MATLABMATLABUAV} \\\hline
         PyFly & Fixed-wing control specific, includes wind and turbulence & No physical environment or rendering & \cite{BoehnJohansen2019Deep}\\\hline
         fixed-wing-gym (with PyFly) & Gym wrapper for PyFly for \ac{RL} applications & No physical environment or rendering & \cite{BoehnJohansen2019Deep}\\\hline
         ARCAD & Fully-actuated \ac{UAV} design, physical interaction tasks & Specific to fully-actuated \acp{UAV} & \cite{KeipourScherer2023UAS} \\\hline
         HIL-airmanip & Human-\ac{UAV} interaction, haptic feedback & Specialized domain, RotorS backend & \cite{cuniato2021hardware} \\\hline
         RotorPy & Comprehensive quadrotor model and aerodynamic effects & Simplistic environments & \cite{FolkKumar2023RotorPy} \\\hline
         RflySim & Model-based design, \ac{SITL} and \ac{HITL}, Unreal, PX4 & Separated into free, full, and enterprise versions & \cite{DaiCai2021RFlySim} \\\hline
         gym-pybullet-drones & Swarming, control and ML focused, aerodynamic effects & Low visual fidelity & \cite{PaneratiSchoellig2021Learning} \\\hline
         QuadSwarm & Swarming, \ac{RL}, sim-to-real transfer & Crazyflie physics model, low visual fidelity & \cite{HuangSukhatme2023QuadSwarm} \\\hline
         Potato & Swarming, data-oriented programming, GPU computation & Not open-source & \cite{LiRen2023Potato} \\\hline
         Agilicious & Agile flight, aerodynamics, open-hardware & Custom license, specialized platform & \cite{FoehnScaramuzza2022Agilicious} \\\hline
         MRS \ac{UAV} System & Onboard vehicle, integration with common simulators & Entire flight stack & \cite{BacaSaska2021MRS} \\\hline
         CrazyChoir & \acs{ROS} 2 framework, distributed swarm control, \ac{SITL} & Crazyflie specific & \cite{PichierriNotarstefano2023CrazyChoir} \\\hline
         Crazyswarm2 & Large indoor swarms, onboard computation, high scalability & Crazyflie specific & \cite{PreissAyanian2017Crazyswarm} \\\hline
         Aerostack2 & Compatibility with various autopilots, Gazebo-based & Does not support (S/H)ITL & \cite{FernandezCortizasCampoy2023Aerostack2} \\\hline
         CrazySim & Swarming SITL, simulated radios, Gazebo-based & Crazyflie specific & \cite{LlanesCoogan2024CrazySim} \\\hline
         sim\_cf2 & Swarming SITL, integrated \acs{ROS} 2 support & Crazyflie specific, built on Gazebo Classic & \cite{simcf2} \\\hline
         X-Plane & Flight simulator, weather, wind, lighting & Commercial, primarily for large-scale aircraft &  \cite{XPlane} \\\hline
         X-PlaneROS (with X-Plane) & \acs{ROS} 1 wrapper, fixed-wings, human-robot interaction & Primarily for large-scale aircraft & \cite{navarro2024sorts} \\\hline
         QPlane & \ac{RL} toolkit for fixed-wings, uses external flight simulator & Primarily for large-scale aircraft & \cite{RichterCalix2021QPlane} \\\hline
         FlightGear & Open-source, user-supported flight simulator & Primarily for large-scale aircraft & \cite{PerryPerry2004flightgear} \\\hline
         RealFlight & RC flight simulator for small multirotors and fixed-wings & Commercial, Windows based  & \cite{RealFlight} \\\hline
    \end{tabular}
    \end{center}
\end{table*}
\FloatBarrier

\begin{table*}[tb]
    \centering
    \caption{Comparison of Features for Widely Used \ac{UAV} Simulators: Included ({\cmark}), Partially Included ({\amark}), and Not Included ({\xmark})}
    \vspace*{-4.5mm}
    \label{tab:sim_features}
    \begin{center}
    \renewcommand{\arraystretch}{1.3}
    \begin{tabular}{C{1.8cm} | C{1.45cm} C{1.3cm} C{0.2cm} C{0.2cm} C{0.2cm} C{1.9cm} C{1.8cm} C{1.4cm} C{0.85cm} C{0.8cm} C{1.0cm}}
        \hline
        \multirow{2}{*}{\centering\textbf{Simulator}} & 
        \multirow{2}{1.45cm}{\centering\textbf{Physics Engine}} & 
        \multirow{2}{*}{\centering\textbf{Rendering}} & 
        \multicolumn{3}{c}{\textbf{OS}} & 
        \multirow{2}{*}{\centering\textbf{Interfaces}} & 
        \multirow{2}{*}{\centering\textbf{(S/H)ITL}} & 
        \multirow{2}{*}{\centering\textbf{License}} & 
        \multirow{2}{0.85cm}{\centering\textbf{Open-Source}} & 
        \multirow{2}{*}{\centering\textbf{Active}} & 
        \multirow{2}{*}{\centering\textbf{Ref.}}
        \\ & & & L & W & M & & & & & & \\ \hline \hline
        Gazebo Classic &  ODE, Bullet, DART, Simbody & OGRE & \cmark & \amark & \cmark & \acs{ROS} 1/2, C++, \ac{RL} & PX4, ArduPilot, CF & Apache-2.0 & \cmark & \cmark & \cite{KoenigHoward2004Design} \\ 
        \arrayrulecolor{gray}\hline
        Gazebo & Bullet, DART, TPE & OGRE & \cmark & \amark & \cmark & \acs{ROS}  1/2, C++, Python, \ac{RL} & PX4, ArduPilot, CF & Apache-2.0 & \cmark & \cmark & \cite{Gazebo} \\ 
        \arrayrulecolor{gray}\hline
        Isaac (Pegasus, Aerial Gym) & NVIDIA PhysX, Flex & Vulkan  & \cmark & \xmark & \xmark & \acs{ROS} 1/2, Python, RL & Pegasus: PX4 & Proprietary (BSD 3) & \xmark~(\cmark, \cmark) & \cmark & \cite{IsaacSim, makoviychuk2021isaac, jacinto2024pegasus, KulkarniAlexis2023Aerial} \\ \arrayrulecolor{gray} \hline
        Webots & ODE & OpenGL & \cmark & \cmark & \cmark & \acs{ROS} 1/2, C/C++, Python, MATLAB, Java & ArduPilot, CF & Apache-2.0 & \cmark & \cmark & \cite{MichelMichel2004Cyberbotics} \\ 
        \arrayrulecolor{gray}\hline
        CoppeliaSim & Bullet, ODE,Vortex, Newton, MuJoCo & OpenGL & \cmark & \cmark & \cmark & \acs{ROS} 1/2, C/C++, Python, MATLAB, Java, Lua, Octave & --- & GNU GPL, Commerical & \amark & \cmark & \cite{RohmerFreese2013CoppeliaSim} \\
        \arrayrulecolor{gray}\hline
        AirSim & NVIDIA PhysX & Unreal, Unity & \cmark & \cmark & \cmark & \acs{ROS} 1, C++, Python, C\#, Java, RL & PX4, ArduPilot & MIT & \cmark & \xmark & \cite{ShahKapoor2018AirSim} \\ 
        \arrayrulecolor{gray}\hline
        Flightmare & Ad hoc, Gazebo Classic & Unity & \cmark & \xmark & \xmark & \acs{ROS} 1, C++, RL & --- & MIT & \cmark & \xmark & \cite{SongScaramuzza2021Flightmare} \\ 
        \arrayrulecolor{gray}\hline
        FlightGoggles & Ad hoc & Unity & \cmark & \amark & \xmark & \acs{ROS} 1, C++ & Motion Capture & MIT & \cmark & \xmark & \cite{GuerraKaraman2019FlightGoggles} \\ 
        \arrayrulecolor{gray}\hline
        gym-pybullet-drones & PyBullet & OpenGL & \cmark & \amark & \cmark & Python, RL & Betaflight, CF & MIT & \cmark & \cmark & \cite{PaneratiSchoellig2021Learning} \\ 
        \arrayrulecolor{gray}\hline
        RotorTM & Ad hoc & OpenGL & \cmark & \xmark & \xmark & \acs{ROS} 1, Python, MATLAB & --- & GNU GPL & \cmark & \cmark & \cite{LiLoianno2023RotorTM} \\ 
        \arrayrulecolor{gray}\hline
        MATLAB \ac{UAV} Toolbox & MATLAB & Unreal & \cmark & \cmark & \cmark & \acs{ROS} 2, MATLAB & PX4 & Proprietary, Commercial & \xmark & \cmark & \cite{MATLABMATLABUAV} \\ \arrayrulecolor{gray} \hline
        \arrayrulecolor{black}\hline
    \end{tabular}
    \end{center}
    \vspace*{-6mm}
\end{table*}

\begin{table*}[tb]
    \centering
    \caption{Comparison of Vehicle Types for Widely Used \ac{UAV} Simulators: Included ({\cmark}), Partially Included ({\amark}), and Not Included ({\xmark})}
    \label{tab:sim_vehicleType}
    \vspace*{-4.5mm}
    \begin{center}
    \renewcommand{\arraystretch}{1.3}
    \begin{tabular}{c | ccc c c c c c c c}
        \hline
        \multirow{2}{*}{\centering\textbf{Simulator}} & \multicolumn{3}{c}{\textbf{Multirotors}} & \multirow{2}{*}{\centering\textbf{Helicopters}} & \multirow{2}{1.2cm}{\centering\textbf{Fixed-wings}} & \multirow{2}{1.2cm}{\centering\textbf{Aerial Manip.}} & \multirow{2}{*}{\centering\textbf{Swarms}} & \multirow{2}{*}{\centering\textbf{Cars}} & \multirow{2}{1cm}{\centering\textbf{Other Systems}} & \multirow{2}{*}{\centering\textbf{Ref.}}\\
         & Basic & Drag & Wind & \\
        \hline \hline
        Gazebo (Classic \& New)  & \cmark & \cmark & \cmark & \xmark & \cmark & \amark & \amark & \cmark & \cmark & \cite{KoenigHoward2004Design, Gazebo} \\ \arrayrulecolor{gray}\hline
        Isaac (Pegasus, Aerial Gym) & \cmark & \xmark~(\cmark, \xmark) & \xmark & \cmark & \xmark & \xmark & \cmark & \cmark~(\xmark, \xmark) & \cmark~(\xmark, \xmark) & \cite{IsaacSim, makoviychuk2021isaac, jacinto2024pegasus, KulkarniAlexis2023Aerial} \\ \arrayrulecolor{gray}\hline
        Webots & \cmark & \xmark & \xmark & \cmark & \xmark & \xmark & \amark & \cmark & \cmark & \cite{MichelMichel2004Cyberbotics} \\ \arrayrulecolor{gray}\hline
        CoppeliaSim & \cmark & \cmark & \amark & \cmark & \xmark & \amark & \amark & \cmark & \cmark & \cite{RohmerFreese2013CoppeliaSim} \\ \arrayrulecolor{gray}\hline
        AirSim & \cmark & \cmark & \cmark & \xmark & \xmark & \xmark & \amark & \cmark & \xmark & \cite{ShahKapoor2018AirSim} \\ \arrayrulecolor{gray}\hline
        Flightmare & \cmark & \cmark & \xmark & \xmark & \xmark & \xmark & \cmark & \xmark & \xmark &  \cite{SongScaramuzza2021Flightmare} \\ \arrayrulecolor{gray}\hline
        FlightGoggles & \cmark & \cmark & \xmark & \xmark & \xmark & \xmark & \xmark & \cmark & \xmark & \cite{GuerraKaraman2019FlightGoggles} \\ \arrayrulecolor{gray}\hline
        gym-pybullet-drones & \cmark & \cmark & \xmark & \xmark & \xmark & \xmark & \cmark & \xmark & \xmark & \cite{PaneratiSchoellig2021Learning} \\ \arrayrulecolor{gray}\hline
        RotorTM & \cmark & \xmark & \xmark & \xmark & \xmark & \cmark & \cmark & \xmark & \xmark & \cite{LiLoianno2023RotorTM} \\ \arrayrulecolor{gray}\hline
        MATLAB \ac{UAV} Toolbox & \cmark & \cmark & \cmark & \xmark & \cmark & \xmark & \amark & \xmark & \xmark & \cite{MATLABMATLABUAV} \\ \arrayrulecolor{gray}\hline
        \arrayrulecolor{black}\hline
    \end{tabular}
    \end{center}
    \vspace*{-8mm}
\end{table*}

\begin{table*}[tb]
    \centering
    \caption{Comparison of Included Sensors for Widely Used Aerial Vehicle Simulators: Included ({\cmark}) and Not Included ({\xmark})}
    \label{tab:sim_sensors}
    \vspace*{-4.5mm}
    \begin{center}
    \renewcommand{\arraystretch}{1.3}
    \setlength{\tabcolsep}{5pt}
    \begin{tabular}{C{3cm} | c c c c c C{0.8cm} C{0.8cm} C{0.8cm} c c c}
    \hline
        \textbf{Simulator} & \textbf{RGB} & \textbf{Depth} & \textbf{Seg.} & \textbf{Point Cloud} & \textbf{IMU} & \textbf{Mag.} & \textbf{GPS} & \textbf{Baro.} & \textbf{LiDAR} & \textbf{Optical Flow} & \textbf{Ref.} \\ \hline 
        \hline
        Gazebo (Classic \& New) & \cmark & \cmark & \cmark & \cmark & \cmark & \cmark & \cmark & \cmark & \cmark & \cmark & \cite{KoenigHoward2004Design, Gazebo} \\ \arrayrulecolor{gray}\hline
        Isaac Sim (Pegasus) & \cmark & \cmark & \cmark & \cmark & \cmark & \xmark~(\cmark) & \xmark~(\cmark) & \xmark~(\cmark) & \cmark & \cmark & \cite{IsaacSim, jacinto2024pegasus} \\ \arrayrulecolor{gray}\hline
        Isaac Gym (Aerial Gym) & \cmark & \cmark & \cmark & \xmark & \xmark & \xmark & \xmark & \xmark & \xmark & \cmark & \cite{makoviychuk2021isaac, KulkarniAlexis2023Aerial} \\ \arrayrulecolor{gray}\hline
        Webots & \cmark & \cmark & \xmark & \xmark & \cmark & \cmark & \cmark & \xmark & \cmark & \xmark & \cite{MichelMichel2004Cyberbotics} \\ \arrayrulecolor{gray}\hline
        CoppeliaSim & \cmark & \cmark & \xmark & \cmark & \cmark & \xmark & \cmark & \xmark & \cmark & \xmark & \cite{RohmerFreese2013CoppeliaSim} \\ \arrayrulecolor{gray}\hline
        AirSim & \cmark & \cmark & \cmark & \cmark & \cmark & \cmark & \cmark & \cmark & \cmark & \cmark & \cite{ShahKapoor2018AirSim} \\ \arrayrulecolor{gray}\hline
        Flightmare & \cmark & \cmark & \cmark & \cmark & \xmark & \xmark & \xmark & \xmark & \xmark & \cmark & \cite{SongScaramuzza2021Flightmare} \\ \arrayrulecolor{gray}\hline
        FlightGoggles & \cmark & \cmark & \cmark & \xmark & \cmark & \xmark & \xmark & \xmark & \xmark & \cmark & \cite{GuerraKaraman2019FlightGoggles} \\ \arrayrulecolor{gray}\hline
        gym-pybullet-drones & \cmark & \cmark & \cmark & \xmark & \xmark & \xmark & \xmark & \xmark & \xmark & \xmark & \cite{PaneratiSchoellig2021Learning} \\ \arrayrulecolor{gray}\hline
       RotorTM & \xmark & \xmark & \xmark & \xmark & \xmark & \xmark & \xmark & \xmark & \xmark & \xmark & \cite{LiLoianno2023RotorTM} \\ \arrayrulecolor{gray}\hline
        MATLAB \ac{UAV} Toolbox & \cmark & \cmark & \cmark & \cmark & \cmark & \xmark & \cmark & \xmark & \cmark & \xmark & \cite{MATLABMATLABUAV} \\ \arrayrulecolor{gray}\hline
    \arrayrulecolor{black}\hline
    \end{tabular}
    \end{center}
    \vspace*{-6mm}
\end{table*}
\noindent indicating the maintenance status at the time of writing this paper. Simulators under active maintenance are marked with \cmark. Simulators that have been inactive but show some commits and responses to issues in the past two years are marked with \amark. Finally, simulators that are intentionally no longer maintained or have been inactive for more than two years are marked with \xmark. As maintenance statuses may change over time, we advise readers to consider this information as a snapshot and to reevaluate before choosing a simulator.

Table~\ref{tab:sim_vehicleType} compares the vehicle types that can be simulated, referencing the dynamics detailed in the ``\ref{sec:background}'' sidebar. 
In the ``Swarms'' column, we specify packages designed for swarm purposes with \cmark, packages that allow multiple vehicles (though not specifically designed for swarms) with \amark, and packages intended solely for single vehicles with \xmark. 
We include columns for cars and other systems (e.g. manipulators, quadrupeds, humanoids) since researchers may require simulating interactions between different platforms with their intended UAV.

Table~\ref{tab:sim_sensors} provides a comparison of supported sensors for each simulator. Segmentation, magnetometer, and barometer are abbreviated as ``Seg,'' ``Mag,'' and ``Baro,'' respectively. 

We indicate features, vehicle types, and sensors supported in the base configurations of these simulators, acknowledging that many of them can be extended for additional support.



\section{Discussion}
\label{sec:discussion}

The aerial robotics community has undeniably made significant strides in the development of simulators. However, a significant challenge lies in the fact that the specific requirements of various research groups tend to be platform-dependent and application-dependent, making it challenging to meet all needs with a single simulator. Moreover, there is a growing consensus that exploring diverse solutions, rather than relying solely on one, can yield more favorable outcomes. Conversely, there is a compelling argument for standardization within this domain, as it would greatly facilitate benchmarking efforts and foster collaboration among researchers. Striking a balance between these two perspectives appears to be the most prudent approach. This entails directing resources and effort towards a select few simulators 
to harness the advantages inherent in both sides of the spectrum mentioned earlier. 
As a result, several key themes emerge, including the role of aerodynamics in simulation, the need for benchmarking, the relationship between academia and industry, data sharing, and the challenges associated with maintaining these simulators.

\textbf{Aerodynamics and Simulation}: One central topic revolves around the role of aerodynamics in \ac{UAV} simulations. 
While it is acknowledged that many successful \ac{UAV} applications do not necessitate intricate aerodynamic modeling, 
in some scenarios such modeling becomes indispensable. 
Particular attention is drawn to scenarios involving \acp{UAV} navigating in constrained environments, adapting to dynamic environmental conditions, or engaging in interactions with other aerial vehicles. In these cases, there is a critical need for incorporating aerodynamics into simulator development, which is even more valid for fixed-wing flight. On the other hand, it is important to note that for most of the applications involving multirotors, even for aerial manipulation tasks, aerodynamics play a less significant role. 

\textbf{Benchmarking and Standardization}: The large number of simulators in existence necessitates benchmarking and standardization in this field. The absence of a unified benchmarking framework and standardization practices pose substantial challenges for researchers and developers alike. Addressing this issue emerges as a primary objective, as it has become evident that standardized benchmarking is paramount for enhancing the reproducibility and comparability of research within the \ac{UAV} and more generally, the robotics domain, as many simulators are designed to be general purpose, such as the universal simulators from Sec.~\ref{sec:universal_sims}.

\textbf{Academia versus Industry}: It is evident from the data reported in the tables that there is a distinction between simulators developed in academic and industrial settings. It is universally acknowledged that academic simulators have played a pivotal role in advancing research. However, 
academic simulators often grapple with sustainability challenges, particularly after 
the researchers responsible for their development, mainly doctoral students, graduate.
In contrast, industry-backed simulators are characterized by robust support, continuous documentation, and sustained evolution. Striking a harmonious balance between academic and industry-driven simulator development emerges as a critical goal for the future.

\textbf{Data Sharing and Collaboration}: What also emerges from the above discussion is the need for data sharing within the research community and with industry partners. It is pivotal to emphasize the importance of sharing simulator data and models for making progress in the field. The potential advantage of shared datasets, particularly in the domains of perception and autonomous navigation, are evident. Collaborative endeavors and knowledge sharing among researchers can catalyze forces for simulator improvements and innovation.

\textbf{Resource Identification}: The numerous simulators discussed in this paper highlight the challenges newcomers face when selecting the most suitable simulator for their research. The demand for resources to streamline the selection process is clear. This guided the authors' choice to write a survey to assist researchers and developers in navigating the diverse landscape of simulators and making well-informed choices.

\textbf{Accessibility and Maintenance}: When selecting a simulator, it is crucial to weigh factors like accessibility and maintainability. The simulator's licensing has a significant impact on its utility and adaptability to research requirements. Simulators lacking open-source availability, a free proprietary license, or an academic license, can pose obstacles to replicating work. We encourage researchers to understand the license restrictions for a chosen simulator. 
Additionally, open-source simulators can be preferable when custom modifications to the simulator's source code are necessary due to missing features. 
Furthermore, evaluating the long-term sustainability of a simulator is essential. A simulator that is no longer actively maintained may become unstable on newer operating systems and may lack updated integration support for middleware like \acs{ROS}. 
This can cause researchers to divert valuable time and effort from their experiments to maintaining the simulator.



\section{Conclusions}
\label{sec:conclusions}

Selecting a simulator that is best for a particular application space can be challenging, but rewarding when it increases safety and reduces testing time and cost. In this paper, we discussed some of the prominent robotic simulators for aerial vehicles. We enumerate possible decision factors to consider when selecting a simulator and we compare features, included vehicle types, and integrated sensors across many widely used simulation packages. 
For researchers new to the field, we recommend starting with a well-supported universal simulator (e.g. Gazebo) and then using this paper to identify specialized solutions as needed. 
We hope that this analysis will be valuable to the community when embarking on aerial vehicle research and selecting a simulation environment.  
Additionally, since the space of simulators is constantly evolving, for updated comparisons, visit our list of the best robot simulators on github: \url{https://github.com/knmcguire/best-of-robot-simulators}.



\section*{Acknowledgment}

We gratefully acknowledge feedback from 
Geoffrey Biggs, Addisu Taddese, Jaeyoung Lim, Jay Patrikar, Marcelo Jacinto, Jo\~{a}o Pinto, Mihir Kulkarni, Marc Freese, Yunlong Song, Jacopo Panerati, Angela Schoellig, and Guanrui Li. 

\addtolength{\textheight}{-8cm}   



\bibliographystyle{IEEEtran}
\bibliography{references}

\end{document}